\documentclass[11pt,fleqn,a4paper]{article}

\newcommand{\nop}[1]{}

\newif\ifmakebbl

\usepackage[german,english]{babel}
\usepackage{setspace}
\usepackage{amsmath}
\usepackage{makeidx}  
\usepackage{latexsym}
\usepackage{graphics}
\usepackage{color}
\usepackage{babel}
\usepackage{epic}
\usepackage{eepic}
\usepackage{amssymb}
\usepackage{supertabular}
\usepackage{times}
\usepackage{theorem}
\usepackage{url}

\nop{***** new format, 
\oddsidemargin 0 pt   
\evensidemargin 0 pt  
\marginparwidth  0 pt 
\topmargin  -0.25in   %
\textheight 8.5in     
\textwidth 6.5in      
****}

\newcounter{myenumctr}

\newenvironment{myenumerate}{\begin{list}{{\bf(\arabic{myenumctr})}}{\usecounter{myenumctr}
\setlength{\leftmargin}{0pt}
\setlength{\itemindent}{\labelwidth}}}
{\end{list}}

\newenvironment{myitemize}{\begin{list}{$\bullet$}{\setlength{\leftmargin}{0pt}
\setlength{\itemindent}{\labelwidth}}}
{\end{list}}

\newcommand{\tenote}[1]{ \\ {\bf *** #1 ***} \\ }

\newtheorem{theorem}{Theorem}[section]
\newtheorem{corollary}[theorem]{Corollary}
\newtheorem{lemma}[theorem]{Lemma}
\newtheorem{proposition}[theorem]{Proposition}
\newtheorem{example}{Example}[section]

\newtheorem{remark}{Remark}[section]
\newtheorem{definition}{Definition}[section]

\newenvironment{proof}
{\emph{Proof:}}{}

\newcommand{\Pol}{{\rm P}}
\newcommand{\PSPACE}{{\rm PSPACE}}
\newcommand{\NP}{{\rm NP}}
\newcommand{\CONP}{{\rm coNP}}
\newcommand{\DP}{{\rm D}^{\rm P}}
\newcommand{\SigmaP}[1]{{\Sigma}_{#1}^{P}}
\newcommand{\PiP}[1]{{\Pi}_{#1}^{P}}

\newcommand{\DeltaPpar}[1]{{\Delta}_{#1}^{P}\|}
\newcommand{\DeltalogP}[1]{{\Delta}_{#1}^{P}[O(\log n)]}

\newcommand{\M}{{\cal M}}
\newcommand{\T}{{\cal T}}
\newcommand{\Lang}{{\cal L}}
\newcommand{\Lcirc}{{\cal L}_{\mathrm{CIRC}}}

\renewcommand{\mod}{\mathit{mod}}

\def\endproof{$\Box$\par\smallskip}

\newcounter{rmexample}

\newcommand{\bigland}{\bigwedge}

\renewcommand{\phi}{\varphi}

\newcommand{\limplies}{\rightarrow}
\newcommand{\lequiv}{\leftrightarrow}

\newcommand{\curb}{\mbox{\small$\mathrm{CURB}$}}
\newcommand{\circum}{\mbox{\small$\mathrm{CIRC}$}}

\title{\vspace*{-\baselineskip}
Complexity of Nested Circumscription and \\ Nested Abnormality Theories%
\thanks{This is a revised and extended version of a preliminary paper
that appeared in: Proc.\ Seventeenth International Joint Conference on
Artificial Intelligence (IJCAI-01), pages 169--174. Morgan Kaufmann,
2001.  }\\[-2ex] ~}

\author{Marco Cadoli\thanks{Dipartimento di Informatica e
Sistemistica, 
Universit{\`a} di Roma ``La Sapienza'', Via Salaria 113, I-00198 Roma,
Italy. Email: {\tt cadoli@dis.uniroma1.it}}
\and Thomas Eiter\thanks{Institut f\"{u}r Informationssysteme, Abtg.\
Wissensbasierte Systeme, Technische Universit\"{a}t Wien, 
Favoritenstrasse~9-11, A-1040 Wien, Austria. Email: {\tt eiter@kr.tuwien.ac.at}}
\and Georg Gottlob\thanks{Institut f\"{u}r Informationssysteme, Abtg.\
Datenbanken und AI, Technische Universit\"{a}t Wien, 
Favoritenstrasse~9-11, A-1040 Wien, Austria. Email: {\tt gottlob@dbai.tuwien.ac.at}}}

\begin{document}

\maketitle

\vspace*{-\baselineskip}

\begin{abstract}
Circumscription has been recognized as an important principle for
knowledge representation and common-sense reasoning. The need for a
circumscriptive formalism that allows for simple yet elegant modular
problem representation has led Lifschitz (AIJ, 1995) to introduce
nested abnormality theories (NATs) as a tool for modular knowledge
representation, tailored for applying circumscription to minimize
exceptional circumstances. Abstracting from this particular objective,
we propose $\Lcirc$, which is an extension of generic propositional
circumscription by allowing propositional combinations and nesting of
circumscriptive theories. As shown, NATs are naturally embedded into
this language, and are in fact of equal expressive capability. We then
analyze the complexity of $\Lcirc$ and NATs, and in particular the
effect of nesting. The latter is found to be a source of complexity,
which climbs the Polynomial Hierarchy as the nesting depth increases
and reaches \PSPACE-completeness in the general case. We also identify
meaningful syntactic fragments of NATs which have lower complexity. In
particular, we show that the generalization of Horn circumscription in
the NAT framework remains \CONP-complete, and that Horn NATs without
fixed letters can be efficiently transformed into an equivalent Horn CNF,
which implies polynomial solvability of principal reasoning
tasks. Finally, we also study extensions of NATs and briefly address
the complexity in the first-order case. Our results give insight into
the ``cost'' of using $\Lcirc$ (resp.\ NATs) as a host language for
expressing other formalisms such as action theories, narratives, or
spatial theories.
\end{abstract}


\section{Introduction}

Circumscription \cite{lifs-85a,mcca-80,mcca-86} is a very powerful
method for knowledge representation and common-sense reasoning, which
has been used for a variety of tasks, including temporal reasoning,
diagnosis, and reasoning in inheritance networks. The basic semantical
notion underlying circumscription is \emph{minimization} of the
extension of selected predicates. This is especially useful when a
predicate is meant to represent an \emph{abnormality} condition, e.g.,
a bird which does not fly.  Circumscription is applied to a formula
$\phi$, either propositional or first-order, and it is used to
eliminate some \emph{unintended} models of $\phi$.

Since the seminal definition of circumscription in \cite{mcca-80},
several extensions have been proposed (see, e.g., Lifschitz's survey
\cite{lifs-94}), all of them retaining the basic idea of
minimization. In this paper, we propose $\Lcirc$, a language which
extends propositional circumscription in two important and rather
natural ways:

\begin{myitemize}
\item on one hand, we allow the \emph{propositional combination} of
  circumscriptive theories;

\item on the other hand, we allow \emph{nesting} of circumscriptions.
\end{myitemize}

As for the former extension, we claim that it can be useful in several
cases. As an example, we consider a scenario from knowledge
integration. Suppose that two different sources of knowledge $\circum(\phi_1)$
and $\circum(\phi_2)$, coming from two equally trustable agents who perform
circumscription, should be integrated. Then, it seems plausible to
take as the result the disjunction of the two sources, i.e., $\circum(\phi_1) \vee
\circum(\phi_2)$.%
\footnote{We remind that $\circum(\phi_1) \vee \circum(\phi_2) \not\equiv
\circum(\phi_1\vee \phi_2)$ in general (take, e.g., $\phi_1 = a \wedge b$
and $\phi_2 = b$).}  In $\Lcirc$, all propositional connectives are
allowed.

As for the latter extension, the concept of \emph{nested abnormality
theories} (NATs) has been proposed by Lifschitz \cite{lifs-95}, in
order to enable a hierarchical application of the circumscription
principle, which supports modularization of a knowledge base and, as
argued, leads sometimes to more economical and elegant formalization
of knowledge representation problems. Since then, NATs have been used
by a number of authors and are gaining popularity as a circumscriptive
knowledge representation tool. For example, NATs have been used in
reasoning about actions
\cite{giun-etal-95,giun-etal-97,kart-lifs-94,kart-lifs-95,son-bara-01},
for handling the qualification problem \cite{kart-01}, formalizing
narratives \cite{bara-etal-98}, expressing function value minimization
\cite{bara-etal-96}, information filtering \cite{bara-00}, describing
action selection in planning \cite{sier-98}, and in spatial reasoning
\cite{remo-kuip-01}.

As another simple example for combining circumscriptions, imagine the task
to diagnose a malfunctioning artifact which is composed of modular
components, e.g., a car. A piece of knowledge $\circum(\phi_1)$ may model
the behavior of a subpart, e.g., the engine, while another one
$\circum(\phi_2)$ may model the behavior of the electrical part, and a plain
propositional formula $\psi$ might encode some observations that are
being made on the car.  Then, by taking the circumscription of a
suitable propositional combination of $\circum(\phi_1)$,
$\circum(\phi_2)$, and $\psi$, unintended models for this scenario may
be eliminated (see Section~\ref{sec:maximizing} for a more concrete
realization of model-based diagnosis).

In this paper, we are mainly concerned with the computational
properties of $\Lcirc$ and NATs, and with the relationships of these
formulas to plain circumscription in this respect. In particular, we
tackle the following questions:

\begin{myitemize}

\item Can NATs be embedded into $\Lcirc$, i.e., is there an
(efficiently) computable mapping from NATs to equivalent $\Lcirc$
formulas? Here, different interpretations of ``equivalence'' are
possible; a strict one requires that $\Lcirc$ formulas and NATs are
built on the same alphabets, and that their models must coincide. A
more liberal one permits the usage of an extended alphabet for $\Lcirc$
formulas, such that the models of a NAT $\T$ correspond to the
projection of the models of its transformation, mapping onto the
original alphabet.

\item What is the precise complexity of reasoning under nested
circum\-scription?  By \emph{reasoning}, we mean both model checking
and formula inference from an $\Lcirc$ formula or a NAT.  Note that methods
for computing certain NATs, by reduction of circumscription axioms to
first-order logic, have been developed; Su \cite{su-95} implemented a
program called CS (Circumscription Simplifier), while Doherty {\em et
al.} came up with their DLS algorithm \cite{dohe-etal-97}, which has
been refined by Gustafsson \cite{gust-96}. However, the precise
complexity of NATs was not addressed in these works.

\item Is there a simple syntactic restriction of NATs (analogously, of
$\Lcirc$) for which some relevant reasoning tasks are not harder than
reasoning in classical logic, or even feasible in polynomial time?
\end{myitemize}

We are able to give a satisfactory answer to all these questions, and
obtain the following main results. 

\begin{myenumerate}
 
\item After providing a formal definition of $\Lcirc$, we prove the
main results about its complexity: model checking and inference are
shown to be \PSPACE-complete (the latter even for literals); moreover,
complexity is proven to increase w.r.t.\ the nesting.  It appears that
nesting, and not propositional combination, is responsible for the
increase in complexity.

\item Similar results are proven for NATs in
Section~\ref{sec:nats}. In this section, we also prove that every NAT
can be easily (and with polynomial effort) translated into a formula
of $\Lcirc$ using auxiliary letters, and thus NATs can be semantically
regarded as a (projective) fragment of $\Lcirc$. By virtue of the
complexity results for NATs, we also provide complexity results for
the corresponding syntactic fragment of $\Lcirc$.

\item Given the high complexity of nested circumscription, we look for
meaningful fragments of the languages in which the complexity is
lower. In this paper, we identify Horn NATs, which are a natural
generalization of Horn circumscriptions, as such fragments.  It is
proven in Section~\ref{sec:horn-nat} that here nesting can be
efficiently eliminated if no fixed variables are allowed, and that
both model checking and inference are polynomial. In particular, we
provide the result that given a Horn NAT $\T$ without fixed letters,
an unnested Horn NAT $\T'$ logically equivalent to $\T$ is
constructible from $\T$ in time linear in the size of the input.

\item Furthermore, we show that also for general Horn NATs (i.e.,
where fixed letters are allowed), model checking is
polynomial. Consequently, inference from a Horn NAT is in \CONP\ (and
thus, by virtue of results on inference from a Horn circumscription in
\cite{cado-lenz-94}, \CONP-complete).  This shows that in general,
nesting does not add to the complexity of Horn NATs. On the other
hand, we show that the use of predicate maximization, proposed in
\cite{lifs-95} as a convenient declaration primitive, increases the
complexity of Horn NATs, which climbs the polynomial hierarchy and
reaches \PSPACE\ if the nesting depth is unlimited.

\item Finally, we compare $\Lcirc$ and NATs to other generalizations of
circumscription, in particular to the well-known method of
\emph{prioritized} circumscription~\cite{lifs-85a,lifs-94} and to
\emph{theory curbing} \cite{eite-etal-93,eite-gott-00}. Prioritized
circumscription can be modeled in a fragment of $\Lcirc$, which has
the same complexity and expressivity as ordinary (unnested)
circumscription. On the other hand, for theory curbing, both model
checking and inference are like for $\Lcirc$ and NATs \PSPACE-complete
\cite{eite-gott-00}. Our main result of the comparison concerns the
\emph{expressiveness} of $\Lcirc$ and NATs, which appears to be lower
than in curbing: in particular, unless some unexpected collapse in
complexity classes occurs, there is no fixed $\Lcirc$ expression that
expresses any \PSPACE-complete problem, while we present a curb
expression of this kind.
\end{myenumerate}

As side results, we provide methods for efficiently eliminating fixed
letters from $\Lcirc$ formulas and from NATs, respectively.

Our results prove that the expressive power that makes $\Lcirc$ and
NATs useful tools for the modularization of knowledge has indeed a
cost, because the complexity of reasoning in such languages is higher
than reasoning in a ``flat'' circumscriptive knowledge base.  Anyway
the \PSPACE\ upper bound of the complexity of reasoning, and the
similarity of their semantics with that of \emph{quantified Boolean
formulas} (QBFs), makes fast prototype implementations possible by
translating them into a QBF and then using one of the several
available solvers, e.g., \cite{rint-99}.
This approach could be used also for implementing meaningful fragments of
NATs, such as the one in \cite{bara-etal-98}, although this might be
inefficient, like using a first-order theorem prover for propositional logic.


Given that QBFs can be polynomially encoded into NATs, we can show
that nested circumscription is more \emph{succinct} than plain
(unnested) circumscription, i.e., by nesting $\circum$ operators (or
NATs), we can express some circumscriptive theories in polynomial
space, while they could be written in exponential space only, if
nesting were not allowed.  In this sense, we add new results to the
{\em comparative linguistics of knowledge
representation}~\cite{gogi-etal-95}.

The rest of this paper is structured as follows. The next section
contains some necessary preliminaries and fixes notation. After this,
we introduce in Section~\ref{sec:language} the language $\Lcirc$,
defining its syntax and semantics, and determine its complexity.  In
Section~\ref{sec:nats}, we then turn to nested abnormality theories;
we show how NATs can be embedded into $\Lcirc$, and by means of this
relationship, we derive the complexity results for the case of general
NATs. In the subsequent Section~\ref{sec:horn-nat}, we then focus our
attention to the syntactic class of Horn
NATs. Section~\ref{sec:further} addresses further issues and presents,
among others, some results for the first-order case and linguistic
extensions to NATs, while Section~\ref{sec:comparison} compares NATs
and $\Lcirc$ to some other generalizations of circumscription, in
particular to prioritized circumscription \nop{\cite{lifs-85a,lifs-94}} and
to curbing\nop{\cite{eite-etal-93,eite-gott-00}}. The final
Section~\ref{sec:conclusion} draws some conclusions and presents open
issues for further work.

\section{Preliminaries}
\label{sec:preliminaries}

We assume a finite set $At$ of propositional atoms, and let
$\Lang(At)$ (for short, $\Lang$, if $At$ does not matter or is clear
from the context) be a standard propositional language over $At$.  An
{\em interpretation} (or {\em model}) $M$ is an assignment of truth values 0
(false) or 1 (true) to all atoms. As usually, we identify $M$ also
with the set of atoms which are true in $M$. The projection of a model
$M$ on a set of atoms $A$ is denoted by $M[A]$. Furthermore, for any
formula $\phi$ and model $M$, we denote by $\phi[M]$ resp.\
$\phi[M[A]]$ the result of substituting in $\phi$ for each atom resp.\
atom from the set $A$ the constant for its truth value.


Satisfaction of a formula $\phi$ by an interpretation $M$, denoted
$M\models \phi$, is defined as usual; we denote by $\mod(\phi)$ the
set of all models of $\phi$. Capitals $P$, $Q$, $Z$ etc stand for
ordered sets of atoms, which we also view as lists. If $X =
\{x_1,\ldots,$ $x_n\}$ and $X'=\{x'_1,\ldots,$ $x'_n\}$, then $X \leq
X'$ denotes the formula $\bigwedge_{i=1}^n(x_i\limplies x'_i)$.

We denote by $\leq_{P;Z}$ the preference relation on models which minimizes $P$ in parallel
while $Z$ is varying and all other atoms are fixed; i.e., $M \leq_{P;Z} M'$
($M$ is more or equally preferable to $M'$) iff $M[P] \subseteq M'[P]$ and
$M[Q]=M'[Q]$, where $Q = At \setminus P\cup Z$ and $\subseteq$ and $=$ are
taken component\-wise. As usual, $M<_{P;Z} M'$ stands for $M\leq_{P;Z}M' \land M\neq
M'$.

We denote by $\circum(\phi;P;Z)$ the second-order circumscription
\cite{lifs-85a} 
of the formula $\phi$ where the atoms in $P$ are minimized, the atoms in
$Z$ float, and all other atoms are fixed, defined as  the
following formula:
\begin{equation}
\circum(\phi;P;Z) =  \phi[P;Z]\land \forall P'Z'((\phi[P';Z']\land
P'\leq P)\limplies P\leq P').
\label{eq:circ}
\end{equation}
Here $P'$ and $Z'$ are lists of fresh atoms (not occurring in $\phi$)
corresponding to $P$ and $Z$, respectively. The second-order formula
(\ref{eq:circ}) is a quantified Boolean formula (QBF) with free
variables, whose semantics is defined in the standard way. Its models,
i.e., assignments to the free variables such that the resulting
sentence is valid, are the models $M$ of $\phi$ which are
\emph{$(P;Z)$-minimal}, where a model $M$ of $\phi$ is $(P;Z)$-minimal, if
no model $M'$ of $\phi$ exists such that $M' <_{P;Z} M$.

\subsection{Complexity classes}

We assume that the reader is familiar with the basic concept
and notions of complexity theory, such as $\Pol$, $\NP$, complete
problems and polynomial-time transformations; for a background, see
\cite{john-90,papa-94}. We shall mainly encounter complexity classes
from the Polynomial Hierarchy (PH), which is contained in \PSPACE. We
recall that $\Pol=\SigmaP{0}=\PiP{0}$, $\NP =\SigmaP{1}$, $\CONP =
\PiP{1}$, $\SigmaP{k+1}=\NP^{\SigmaP{k}}$, and
$\PiP{k}=\mbox{co-}\SigmaP{k}$, $k\geq 1$, are major classes in
PH. The class $\DP_k = \{ L\,{\times}\, L' \mid L\in \SigmaP{k}, L'\in
\PiP{k}\}$, $k\geq 0$, is the ``conjunction'' of $\SigmaP{k}$ and
$\PiP{k}$; in particular, $\DP_1$ is the familiar class $\DP$.  All
the classes with $k\geq 1$ have complete problems under
polynomial-time transformations, and canonical ones in terms of
evaluating formulas from certain classes of QBFs.
The problems in the class $\DeltalogP{k+1}$ are those which can be solved in 
polynomial time with $O(\log n)$ many calls to an
oracle for $\SigmaP{k}$, where $n$ is the input size.

A complexity class $C$ is called {\em closed under polynomial
conjunctive reductions}, if the existence of any polynomial-time
transformation of problem $A$ into a logical conjunction of
(polynomially many) instances of a fixed set of problems
$A_1$,\ldots,$A_l$ in $C$ in implies that $A$ belongs to $C$. Note
that many common complexity classes are closed under polynomial
conjunctive reductions. In particular, it is easily seen that this
holds for all complexity classes mentioned above.

\section{Language $\Lcirc$}
\label{sec:language}

The language $\Lcirc$ extends the standard propositional language
$\Lang$ (over a set of atoms $At$) by circumscriptive atoms.

\begin{definition}
Formulas of $\Lcirc$ are inductively built as follows: 
\begin{enumerate}
\item $a\in \Lcirc$, for every $a \in At$;
\item if $\phi$, $\psi$ are in $\Lcirc$, then $\phi\land\psi$ and
   $\neg\phi$ are in $\Lcirc$;
    \item if $\phi\in \Lcirc$ and $P,Z$ are disjoint lists of atoms,
        then $\circum(\phi;P;Z)$ is in $\Lcirc$ (called {\em
        circumscriptive atom}).
\end{enumerate}
\end{definition}
Further Boolean connectives ($\lor$, $\rightarrow$, etc) are defined
as usual. The semantics of any formula $\phi$ from $\Lcirc$ is given
in terms of models of a naturally associated QBF $\tau(\phi)$, which
is inductively defined as follows:
\begin{enumerate}
  \item $\tau(a) = a$, for any atom $a\in At$;
  \item $\tau(\phi\land\psi) = \tau(\phi)\land\tau(\psi)$;
  \item $\tau(\neg\phi) = \neg\tau(\phi)$; and
  \item\label{circ-at}$\tau(\circum(\phi;P;Z)) \,=\, \tau(\phi[P;Z])\land \forall
  P'Z'((\tau(\phi[P';Z'])\land P'\!\leq\!P) \limplies P\!\leq\!P')$.
\end{enumerate}

Note that in \ref{circ-at}, the second-order definition of
circumscription is used to map the circumscriptive atom to a QBF which
generalizes the circumscription formula in (\ref{eq:circ}).  In
particular, if $\phi$ is an ordinary propositional formula ($\phi
\in \Lang$), then $\tau(\circum(\phi;P;Z))$ coincides with the formula in
(\ref{eq:circ}). Furthermore, observe that $\Lcirc$ permits
replacement by equivalence, i.e., if $\psi_1$ and $\psi_2$ are
logically equivalent formulas from $\Lcirc$ and $\psi_1$  occurs in
formula $\phi$, then any formula resulting from $\phi$ by replacing
arbitrary occurrences of $\psi_1$ in $\phi$ by $\psi_2$ is logically
equivalent to $\phi$. 

\begin{example}\label{exa:lcirc}
{\rm Consider the formula
$$\phi = \circum( \circum(a \vee b; a; b) \vee \circum(b \vee c; b; c); a;
c).
$$
Since $\circum(a \vee b; a; b) \equiv (b \wedge \neg a)$ and 
$\circum(b \vee c; b; c) \equiv (c \wedge \neg b)$, we get 
$$
\tau(\phi) \equiv \circum( (b \wedge \neg a) \vee (c \wedge \neg b); a;c).
$$
 From rule 4, we get by applying ordinary circumscription that }
$$
\tau(\phi) \equiv (\neg a \land \neg b\land c) \lor (\neg a\land b) \equiv \neg
a \land (b \lor c).
$$

\end{example}

\vspace*{-\baselineskip}

\noindent\endproof

As usual, we write $M\models \phi$ if $M$ is a model of $\phi$ (i.e.,
$M$ satisfies $\phi$), and $\phi \models \psi$ if $\psi$ is a logical
consequence of $\phi$, for any formulas $\phi$ and $\psi$ from $\Lcirc$.

\subsection{Complexity results}
\label{sec:lcirc-comp}

Let the {\em $\circum$-nesting depth} (for short, nesting depth) of
$\phi \in \Lcirc$, denoted $nd(\phi)$, be the maximum number of
circumscriptive atoms along any path in the formula tree of $\phi$.

\begin{theorem}
\label{theo:lcirc-mc}
Model checking for $\Lcirc$, i.e., deciding whether a given
interpretation $M$ is a model of a given formula $\phi \in \Lcirc$, is
\PSPACE-complete. If $nd(\phi) \leq k$ for a constant $k> 0$, then the
problem is (i) $\PiP{k}$-complete, if $\phi$ is a circumscriptive atom
$\circum(\psi;P;Z)$, and (ii) $\DeltalogP{k+1}$-complete in general.
\end{theorem}

\begin{proof}
By an inductive argument, we can see that for any circumscriptive atom
$\phi = \circum(\psi;P;$ $Z)$ such that $nd(\phi)\leq k$ for constant
$k$, deciding $M\models \phi$ is in $\PiP{k}$. Indeed, if $k=1$, then
$\phi$ is an ordinary circumscription, for which deciding $M\models
\psi$ is well-known to be in $\CONP=\PiP{1}$, cf.\
\cite{eite-gott-93e}.  Assume the statement holds for $k\geq 1$, and
consider $k'=k+1$.
Note that $M\not\models\phi$ iff
either (a) $M\not\models\psi$ or (b) some model $N$ exists such that
$N\!<_{P;Z}\!M$ and $N\models \psi$. By the induction hypothesis, we can
guess $N$ and check whether either (a) or (b) holds for this $N$ in
polynomial time using a $\PiP{k}$ oracle. It follows that deciding
$M\models \phi$ is in $\PiP{k+1}$, as claimed. This establishes the
membership part for (i). If $nd(\phi)=k$ but $k$ is not fixed, we
obtain similarly that deciding $M\models \phi$ is possible by a
recursive algorithm, whose nesting depth is bounded by $nd(\phi)$ and
which cycles through all possible candidates $N$ for refuting $M$, in
quadratic space. Since any $\Lcirc$ formula $\phi$ is equivalent to
the circumscriptive atom $\circum(\phi;\emptyset;\emptyset)$, deciding
$M\models \phi$ is thus in \PSPACE\ in general.

For the membership part of (ii), observe that $\phi$ is a Boolean combination of ordinary and
circumscriptive atoms $\phi_1,\ldots,\phi_m$ such that
$nd(\phi_i))\leq k$ holds for $i\in\{1,\ldots,m\}$. Deciding $M\models
\phi$ is easy if the values of all $\phi_i$ in $M$ are known; by (i),
they can be determined in parallel with calls to $\PiP{k}$
oracles. Thus, deciding $M\models \phi$ is possible in
$\DeltaPpar{k+1}$, i.e., in polynomial time with one round of parallel
$\SigmaP{k}$ oracle calls.  Since, as well-known,
$\DeltaPpar{k+1}=\DeltalogP{k+1}$ (see \cite{wagn-90} for $k=1$, which
easily generalizes), this proves the membership part for (ii).

\PSPACE-hardness of deciding $M\models \phi$ for general $\phi$ and
$\PiP{k}$-hardness for (i) can be shown by a reduction from evaluating
suitable prenex QBFs.  We exploit that nested abnormality theories
(NATs) can be easily embedded into $\Lcirc$ in polynomial time (cf.\
Proposition~\ref{prop:sigmaprime}), and thus a slight adaptation of
the reduction of QBFs to model checking for NATs in the proof of
Theorem~\ref{theo:nat-mc} proves those hardness results. In
particular, we perform the reduction there for empty $X_{n+1}$ (the
formulas $\phi_g$ and $\phi_c$, which become tautologies, can be
removed), and observe that in this case, each auxiliary letter $p \in
A^*(\T)$ is uniquely defined by some formula $u\lequiv p$ or $u\lequiv
\neg p$, respectively, in some $\T'_j$. Thus, the problem $M\models
\T'_n$ in the proof of Theorem~\ref{theo:nat-mc} can be reduced, for
empty $X_{n+1}$, in polynomial time to an equivalent model checking
problem $M^*\models \sigma^{\star}(\T'_n)$ for $\Lcirc$. It follows
that model checking for $\Lcirc$ is \PSPACE-hard in general and
$\PiP{k}$-hard in case (i).

The $\DeltalogP{k+1}$-hardness part for the case where $\phi$ is a
Boolean combination of formulas $\phi_1,\ldots,\phi_m \in \Lcirc$ such
that $\max\{ nd(\phi_i)\mid i\in\{1,\ldots,m\}\} \leq k$ is then shown by a reduction
from
 the problem of
deciding, given $m$ instances
$(M_1,\phi_1)$,\ldots,$(M_n,\phi_m)$
of the model checking problem for circumscriptive atoms on disjoint
alphabets $At_1,\ldots,At_m$, respectively, whether the number of
yes-instances among them is even. The $\DeltalogP{k+1}$-completeness
of this problem is an instance of Wagner's \cite{wagn-90} general
result for all $\PiP{k}$-complete problems. Moreover, we may assume
that $m$ is even
and use the assertion (cf.\ \cite{wagn-90}) that
$(M_i,\phi_i)$ is a yes-instance only if $(M_{i+1},\phi_{i+1})$ is a
yes-instance, for all $i\in \{1,\ldots,m-1\}$. Then, we can define
$$
\phi = e \lequiv \Big(\phi_1 \lor \bigvee_{0< 2i < m}(\neg
\phi_{2i} \,\land\, \phi_{2i+1})\lor \neg\phi_m\Big), 
$$
where $e$ is a fresh letter. The interpretation $M =
\bigcup_{i=1}^m M_i \cup \{ e \}$ is a model of $\phi$ if and only if
the number of yes-instances among $(M_1,\phi_1)$,\ldots,
$(M_m,\phi_m)$ is even. Clearly, $\phi$ and $M$ can be constructed in
polynomial time.
\end{proof}

\begin{theorem}
Deciding, given formulas $\phi, \psi \in \Lcirc$ whether $\phi \models
\psi$ is \PSPACE-complete. Hardness holds even if $\psi \in
\Lang$.  If the nesting depth of $\phi$ and $\psi$ is bounded by the constant
$k\geq 0$, then the problem is $\PiP{k+1}$-complete.
\end{theorem}

\begin{proof} The problem is in \PSPACE\ (resp., $\PiP{k+1}$): By Theorem~\ref{theo:lcirc-mc},
an interpretation $M$ such that $M\models
\phi\land \neg\psi$ can be guessed and verified in
polynomial space (resp., in $\DeltalogP{k+1}$, thus in polynomial time
with an oracle for $\PiP{k}$). Hence the problem is in NPSPACE =
\PSPACE\ (resp., in $\PiP{k+1}$). Hardness follows from the polynomial
time embedding of NATs into $\Lcirc$ (Corollary~\ref{coroll:embed})
and Theorem~\ref{theo:nat-inference} below.%
\nop{ *** OPTIONAL; SHORTEN or rewrite ***

For the hardness part, of the general case, we just sketch a proof
here (it follows also from the \PSPACE-hardness of NATs, which are
polynomially embedded in $\Lcirc$). We take any QBF
$$
\Phi = Q_n X_n Q_{n-1}X_{n-1}\cdots \forall X_2 \exists X_1 E,
$$
where the quantifiers $Q_i$ alternate. Let 
$\phi_0 = (E \lor (u\land X_1))\land (u \lequiv \neg u')$,
where $u$  and $u'$ are fresh atoms. \tenote{GEORG: DELETE $X_1$ from
$\phi_0$} Then we define: 
\begin{eqnarray*}
\phi_1 &=& \circum(\phi_0;\, u ;\, X_1\cup \{ u'\}), \\
\phi_2 &=& \circum(\phi_1;\, u' ;\, X_1\cup X_2 \cup \{ u\}), \\
\phi_{2i+1} &=& \circum(\phi_{2i};\, u ;\, X_1\cup \cdots\cup X_{2i+1}\cup \{ u'\}), \\
\phi_{2i} &=& \circum(\phi_{2i-1};\, u' ;\, X_1\cup \cdots\cup X_{2i}\cup \{ u\}), \\
\end{eqnarray*}
for $0\leq 2i,2i+1<n$.
Then, we have the following: 
\begin{itemize}
 \item If $n=2k$ is even, then $\phi_{n-1} \models u$ iff $\Phi$  is
 \tenote{GG: $u \rightarrow \neg u$ or $u'$} true;
 \item If $n=2k+1$ is odd, then $\phi_{n-1} \models \neg u$ iff $\neg\Phi$
       is true, i.e., $\Phi$ is false;  \tenote{GG: $\neg u \rightarrow u$}
\end{itemize}
*** END OPTIONAL ***
}
\end{proof}

As an immediate corollary, we obtain the following results for the
satisfiability in $\Lcirc$. 

\begin{corollary}
Deciding satisfiability of a given formula $\phi \in \Lcirc$ is
\PSPACE-complete. If the nesting depth is bounded by a constant
$k\geq 0$, then the problem is $\SigmaP{k+1}$-complete. 
\end{corollary}

Observe that some of the hardness proofs in this section make use of
results from Section~\ref{sec:nats}. In turn, the membership results
for reasoning problems in $\Lcirc$ will be convenient to establish
membership results for some of the problems considered there.

\section{Nested Abnormality Theories (NATs)}
\label{sec:nats}

In this section, we turn to Lifschitz's \cite{lifs-95} formalization
of nested circumscription, which we introduce here in the
propositional setting (see Section~\ref{sec:fo} for the predicate
logic context).

We assume that the atoms $At$ include a set of distinguished atoms $Ab
= \{ ab_1,\ldots,ab_k\}$ (which intuitively represent abnormality
properties). 

\begin{definition} \emph{Blocks} are defined as the smallest set such that
if $c_1,\ldots, c_n$ are distinct atoms not in $Ab$, and each of
$B_1,\ldots,B_m$ is either a formula in $\Lang$ or a block, then
$$
B=\{ c_1,\ldots, c_n : B_1,\ldots,B_m\},
$$
is a block, where $c_1,\ldots,c_n$ are called
{\em described} by this block. The {\em nesting
depth} of $B$, denoted $nd(B)$, is 0 if every $B_i$ is from the
language $\Lang$,
and $1 + \max\{nd(B_i) \mid 1 \leq i\leq m\}$ otherwise.
\end{definition}

\begin{definition}
\label{def:nat}
A {\em nested abnormality theory} (NAT) is a collection
$\T = B_1,\ldots,B_n$ of blocks;%
\footnote{In \cite{lifs-95}, the collection may be infinite. For our
concerns, only finite collections are of interest.}
its nesting depth, denoted $nd(\T)$,
is defined by $nd(\T) = \max\{nd(B_i) \mid 1 \leq i\leq n\}$.
\end{definition}

\begin{example}\label{exa:nat}
{\rm This is a propositional version of the example in section 3.1 of
\cite{lifs-95}.  $\T$ is the following NAT with two blocks:
\[ \{f:~~~ f \rightarrow ab, ~~~B\},
\]
where block $B$ is defined as:
\[ \{f:~~~ b \wedge \neg ab \rightarrow f,~~~ c\rightarrow b,~~~ c \}.
\]
Letters $f$, $b$, and $c$ stand for ``flies'', ``bird'', and ``canary'',
respectively. The outer block describes the ability of objects to fly; the
inner block $B$ gives more specific information about the ability of birds to
fly.\hfill$\Box$}
\end{example}

The semantics of a NAT $\T$ is defined
by a mapping $\sigma(\T)$ to a QBF as follows:
\begin{equation}
\label{sigma-nat}
\sigma(\T) = \bigwedge_{B \in \T} \sigma(B),
\end{equation}
 where for any block $B = \{ C : B_1,\ldots,B_m \}$, 
\begin{equation}
\label{sigma-block}
\textstyle \sigma(B) =  \exists Ab.\circum\Big(\bigwedge_{i=1}^m\sigma(B_i);Ab;C\Big)
\end{equation}
given that $\sigma(\phi) = \phi$ for any formula $\phi \in
\Lang$. Satisfaction of a block $B$ (resp., NAT $\T$) in a model $M$
is denoted by $M\models B$ (resp., $M\models \T$). 

A standard circumscription $\circum(\phi;P;Z)$, where $\phi \in
\Lang$, is equivalent to a NAT $\T = \{ Z: \phi\}$ where $P$ is viewed
as the set of abnormality letters $Ab$; notice that
$nd(\T)=0$. However, in this expression, the letters $P$ are projected
from the models of $\T$.  Furthermore, any ordinary formula $\phi\in
\Lang(At\setminus Ab)$ is logically equivalent to the NAT
$\{~:\phi\}$.

\begin{remark}
{\rm By our definitions, a model $M$ of a block $B$ comprises all
letters, $At$, including $Ab$, which is not
the case according to \cite{lifs-95}. More rigorously, we
would need to use abnormality letters as 0-ary predicate (i.e.,
propositional) variables and distinguish them from the other letters,
which are 0-ary predicate constants. For the
purpose of this paper, it simplifies the discussion to have models of
blocks and NATs on an alphabet which has $Ab$ also constants; our results are not affected
by this in essence.  Note that $M$ can take any value on $Ab$ for $B$,
since by $\sigma(B)$ as in (\ref{sigma-block}), the valuation of $Ab$
as a variable in $\exists Ab$ is locally defined and projected away
via the quantifier.}
\end{remark}

For later use, we note the following simple characterization of the
models of a block.

\begin{proposition}
\label{prop:minimal-char}
Let $M$ be an interpretation of all letters in $At$
and $B=\{ C: B_1,\ldots, B_m\}$ a block. Then
$M\models B$ if and only if there exists a model $M^*$ which extends
$M[At\setminus Ab]$ (i.e., $M[At\setminus Ab] = M^*[At\setminus Ab]$)
and is a $(Ab;C)$-minimal model of $B_1$, \ldots, $B_m$.
\end{proposition}

We call any model $M^*$ as in the previous proposition a {\em witness
extension of $M$} (w.r.t.\ $B$); if $M$ is a witness extension of
itself (i.e., $M=M^*$), then we call $M$ a {\em witness model of
$B$}. Thus, $M$ is a witness model of $B$ precisely if $M \models
\circum\Big(\bigwedge_{i=1}^n\sigma(B_i);Ab;C\Big)$ holds.

\medskip
\noindent
{\bf Example~\ref{exa:nat} (cont.)}
The semantics $\sigma(\T)$ of $\T$ can be easily obtained using the above
definition:
\begin{eqnarray*}
  \sigma(\T) & = &
  \sigma(\{f:~ f \rightarrow ab\}) ~ \wedge ~ \sigma(\{f:~ B\})\\
  & = & f \rightarrow ab ~ \wedge ~
  \exists Ab.\circum(b \wedge \neg ab \rightarrow f,~ c\rightarrow b,~ c; ab;
  f)\\
  & = & f \rightarrow ab ~ \wedge ~
  \exists Ab.\circum(c \wedge b \wedge (f\vee ab); ab;  f)\\
  & = & f \rightarrow ab ~ \wedge ~
  \exists Ab.(c \wedge b \wedge f\wedge\neg ab)\\
  & = & f \rightarrow ab ~ \wedge ~(c \wedge b \wedge f)\\
  & = & f \wedge ab \wedge c \wedge b.
\end{eqnarray*}
Note that $\{c , b , f, \neg ab\}$ is a witness model of $B$.
\hfill$\Box$

\medskip
\noindent
The following useful proposition states that we can easily group
multiple blocks into a single one.
\begin{proposition}
\label{prop:single-block}
Let $\T = B_1,\ldots,B_n$ be any NAT. Let  $\T' = \{ Z : B_1,\ldots, B_n\}$ where
$Z$ is any subset of the atoms (disjoint with $Ab$). Then, $\T$ and $\T'$ have the same
models. 
\end{proposition}

Indeed, $\T'$ has void minimization of $Ab$ (making each $ab_j$ in
$Ab$ false), and fixed and floating letters can have any values.

\subsection{Embedding NATs into $\Lcirc$}

In the translation $\sigma(\T)$, the minimized letters $Ab$ are under
an existential quantifier, and thus semantically ``projected'' from
the models of the formula $\circum(\cdots)$ (recall that $Ab$, which is by our
convention respected by models of $\sigma(\T)$, has arbitrary value in
them.) We can, modulo abnormality and auxiliary letters,
eliminate the existential quantifiers from the NAT formula
$\sigma(\T)$ as follows.

\begin{definition} Let, for any NAT $\T$, be $\sigma^{\star}(\T)$ the
formula obtained from  $\sigma(\T)$ as follows:
\begin{enumerate}
  \item Rename every quantifier $\exists Ab$ in $\sigma(\T)$ such
that every quantified variable is different from every other variable. 

\item In every circumscriptive subformula $\circum(\phi;P;Z)$ of the
renamed formula, add
to the floating atoms all variables which are quantified in $\phi$
(including in its subformulas). 
\item Drop all quantifiers. Let $A^*(\T)$ denote the set of all
variables whose quantifier was dropped.
\end{enumerate}
\end{definition}

Note that the size of $\sigma^{\star}(\T)$ is polynomial (more
precisely, quadratic) in the size of $\sigma(\T)$, and also quadratic
in the size of $\T$.

\begin{example}\label{exa:sigmaprime}
{\rm Let $Ab=\{ab_1,ab_2\}$ and $\T=\{ z:\T_1,\, ab_1\lequiv z\}$, where  $\T_1 = \{ z : ab_1 \lequiv \neg ab_2,\, ab_1
\lequiv z\}$. Then,
\begin{eqnarray*}
\sigma(\T) &=&\exists ab_1,ab_2.\circum(\sigma(\T_1)\land
(ab_1\!\lequiv\!z);ab_1,ab_2; z),  \textrm{ where } \\
\sigma(\T_1)&=&\exists ab_1,ab_2.\circum((ab_1\lequiv \neg ab_2)\land (ab_1 \lequiv z);ab_1,ab_2; z).
\end{eqnarray*}
In Step~1, we rename $ab_1$ and $ab_2$ in $\sigma(\T_1)$ to $ab_3$ and
$ab_4$, respectively, and add in Step~2 
$ab_3,ab_4$ to the floating letter $z$ of $\T$. After
dropping quantifiers in Step~3, we obtain: 
\begin{eqnarray*}
\sigma^{\star}(\T) &=& \circum(\sigma^{\star}(\T_1)\land (ab_1\lequiv z);\, ab_1,ab_2;
z,ab_3,ab_4),\\
\sigma^{\star}(\T_1) &=&\circum((ab_3\lequiv \neg ab_4)\land (ab_3
\lequiv z);\, ab_3,ab_4; z).
\end{eqnarray*}
Furthermore, $A^*(\T) = \{ ab_1, ab_2, ab_3, ab_4 \}$. 
\hfill$\Box$}
\end{example}


The following result
states the correctness of $\sigma^{\star}$.

\begin{proposition}
\label{prop:sigmaprime}
For any NAT $\T$, $\sigma(\T)$ and  $\sigma^{\star}(\T)$ are
logically equivalent modulo $A^*(\T)$. Moreover, if $\T$ is a
single block $B$ and renaming takes place inside the nesting, then an
interpretation $M$ of $At$ is a witness model of $\T$ if and only if
$M=N[At]$ for some model $N$ of $\sigma^{\star}(\T)$.
\end{proposition}

\begin{proof}
We prove the result for any
$\T$ which is a single block $B$ by induction on $k\geq 0$ given
$nd(\T)\leq k$. The equivalence result for arbitrary $\T$ follows then from
Proposition~\ref{prop:single-block}. In what follows, we use the
obvious fact that the 
models and the witness models of $B$ coincide modulo $Ab$. 

(Basis) If $k=0$, then $\sigma(\T)$ is an ordinary circumscription
$\exists Ab.\circum(\phi;Ab;Z)$ where $\phi \in \Lang$. Clearly, every
witness model $M$ of $\sigma(\T)$ (in the alphabet $At$) is, modulo
possible renamings of letters from $Ab$ in $\sigma^{\star}(\T)$, a
model of $\sigma^{\star}(\T)$ (in the alphabet $(At\setminus Ab) \cup
A^*(\T))$, and vice versa. Thus the statements hold in this case.

(Induction) Assume the statements hold for $k\geq 0$. Let $\T$ be
a single block $B = \{ Z : B_1,$ $\ldots,$ $B_n\}$ of nesting
depth $nd(B)=k+1$. Then, $\sigma(\T) = \exists Ab.\circum(\phi;Ab;Z)$
where $\phi=\bigwedge_i\sigma(B_i)$. Suppose $\sigma^{\star}(\T) =
\circum(\phi';Ab';Z')$ where $\phi' =\bigwedge_i\sigma^{\star}(B_i)$,
such that, without loss of generality, $Ab'=Ab$ (i.e., renaming in Step 1 of
$\sigma^{\star}(\T)$ takes place inside the nesting) and
$B_1,\ldots,B_l$ ($l\leq n$) are all the blocks $B_i$ in $B$ such that
$B_i\in \Lang$. Note that $Z' =
Z \cup$ $\bigcup_{i=l+1}^n Ab_i$, where $Ab_i$ are the abnormality
letters in $\sigma^{\star}(B_i)$; note that the sets $Ab_{l+1}$,\ldots, $Ab_n$ and
$Ab$ are pairwise disjoint.

Let $M$ be any witness model of $B$, i.e., $M\models
\circum(\phi;Ab;Z)$. We show that $\sigma^{\star}(\T)$ has a model $N$
such that $M = N[At]$.  Since $M\models \phi$, we have $M\models \sigma(B_i)$, for
$i\in\{1,\ldots,n\}$. Thus, $M\models \sigma^{\star}(B_i)$, if $i\leq
l$, since $\sigma^{\star}(B_i)=\sigma(B_i)$ (=$B_i)$.
For $i>l$, $\sigma(B_i)$ is of the form $\exists
Ab.\circum(\phi_i;Ab;Z_i)$. By the
induction hypothesis, there is a truth assignment $\nu_i$ to
$Ab^\star_i$  such that the extension of $M$ to $Ab^\star_i$ by $\nu_i$
is a model of $\sigma^{\star}(B_i) = \circum(\phi'_i;Ab'_i;Z'_i)$. 
Since the sets $Ab^\star_{l+1},\ldots,Ab^\star_n$ and $Ab$ are
pairwise disjoint, the extension of $M$ to $\bigcup_{i=l+1}^n Ab^\star_i$ by
$\nu_{l+1}$,\ldots, $\nu_n$, denoted $N$, is therefore a model of
$\phi'$. Furthermore, 
it holds that $N \models \circum(\phi';Ab';Z')$. Indeed, assume towards
a contradiction that some
model $N'$ of $\phi'$ exists such that $N' <_{Ab';Z'} N$. Then, projected to the
letters of $\sigma^{\star}(B_i)$, $N'$ is a model of
$\sigma^{\star}(B_i)$, for each $i\in \{1,\ldots,n\}$. The induction
hypothesis implies that $M' :=N'[At]$ 
is a model of each $\sigma(B_i)$, and thus $M' \models \phi$. Since
$Ab=Ab'$, we have $M' <_{Ab;Z} M$, and thus $M$ is not an $(Ab;Z)$-minimal model
of $\phi$. This contradicts that $M$ is a witness model of
$B$. Consequently, $N\models \circum(\phi';Ab';Z')$. Thus, $N$ is a model of $\sigma^{\star}(\T)$
such that $M=N[At]$.

Conversely, let $N$ be a model of $\sigma^{\star}(\T)$. Then, for each
$i \in \{ 1,\ldots,n\}$, the projection of $N$ to the letters for $\sigma^{\star}(B_i)$, denoted $N_i$, is a model of
$\sigma^{\star}(B_i)$. Thus, $N_i[At]$ if $i\leq l$, and, as follows
from the induction hypothesis, $N_i[At\setminus Ab^\star_i]$ if $i > l$ 
is a model of $\sigma(B_i)$. Hence, $M := N[At]$ is a model of $\phi$. Moreover, $M$ is an
$(Ab;Z)$-minimal model of $\phi$. Indeed, suppose that $M' <_{Ab;Z} M$
is a smaller model of $\phi$. Since $M' \models \sigma(B_i)$, for
$i\in \{1,$ $\ldots,$ $n\}$, we have $M'\models \sigma^{\star}(B_i)$
if $i\leq l$ and, by the induction hypothesis, there exists an extension $N'_i$ of
$M'_i[At\setminus Ab]$ to $Ab^\star_i$ such that $N'_i \models
\sigma^{\star}(B_i)$, for each $i\in \{l+1,\ldots, n\}$. Since the
sets $A^\star_{l+1},\ldots,A^\star_n$ and $Ab$ are pairwise disjoint,
$N' = M' \cup \bigcup_{i=l+1}^n N_i$ extends $M'$ to
$\bigcup_{i=l+1}^n Ab^\star_i$ such that $N' \models \phi'$ and $N' <_{Ab;Z'} N$. This implies that $N$ is not a model of
$\sigma^{\star}(\T)$, which is a contradiction. This shows that $M$
is an $(Ab;Z)$-minimal model of $\phi$. Consequently, $M$ is a witness model of $\T$.

Thus, the statements hold for $k+1$, which concludes the induction.
\end{proof}

\begin{corollary}
\label{coroll:embed}
Modulo the letters $A^*(\T)$, NATs are (semantically) a fragment of
$\Lcirc$, and poly\-nomial-time embedded into $\Lcirc$ via $\sigma^{\star}$. 
\end{corollary}

We remark that auxiliary letters seem indispensable for an efficient
embedding of NAT into $\Lcirc$; intuitively, they are needed in
compensation for repetitive local use of projected abnormality
letters.  Notice that it is not possible to add in Step~2 of the
embedding $\sigma^{\star}(\T)$ the quantified variables in $\phi$ to
the fixed atoms. This is shown by the following example.

\begin{example}\label{exa:sigmaprime-cont} {\rm Reconsider the NAT $\T$ in
Ex.~\ref{exa:sigmaprime}. Note that $\emptyset$ is the unique model of
$\sigma(\T)$. The formula $\sigma^{\star}(\T_1)$ has, if we disregard
$ab_1, ab_2$ (which are fixed in it), the models $M_1=\{ ab_3, z\}$
and $M_2=\{ ab_4 \}$. They give rise to the two models $N_1=\{ ab_1,
z, ab_3 \}$ and $N_2=\{ ab_4\}$ of $\sigma^{\star}(\T_1)\land
(ab_1\lequiv z)$, of which $N_2$ is $(ab_1,ab_2;ab_3,ab_4,z)$-minimal.

However, if $ab_3$, $ab_4$ were fixed in $\sigma^{\star}(\T)$, 
then both $N_1$ and $N_2$ would be models of $\sigma^{\star}(\T)$,
as they are $(ab_1,ab_2;z)$-minimal. Therefore,
Proposition~\ref{prop:sigmaprime} would fail.} \hfill\endproof
\end{example}

We finally remark that $\Lcirc$ formulas can be embedded, modulo
auxiliary letters, into equivalent NATs in polynomial time.  This can
be seen from the fact that $\Lcirc$ formulas can be embedded into QBFs
(having free variables) in polynomial time, and that such QBFs can be
embedded, using auxiliary letters, into NATs in polynomial time (cf.\
also the next section). However, by the limited set of constructors in
NATs, and in particular the lack of negation applied to blocks, a
simple and appealing polynomial-time embedding of $\Lcirc$ into NATs seems not
straightforward.

\subsection{Complexity of NATs}
\label{sec:nat-comp}

Ordinary circumscription can express a QBF sentence $\Phi=\forall
X\exists Y \psi$ (where $\psi \in \Lang$) as follows. Let $u$ be a
fresh atom.

\begin{proposition}[cf.\ \cite{eite-gott-93e}]
\label{lem:tcs}
$\Phi$ is true if and only if $\circum(\phi;u;Y)\models \neg u$, where $\phi
= \psi \lor u$.
\end{proposition}


This circumscription can be easily stated as a NAT. Set 
$$
\T_1= \{ Y, u : \phi, u \lequiv ab\}.
$$
Then Proposition~\ref{lem:tcs} implies that $\T_1 \models \neg u$ iff $\Phi$
is true. Recall that $M[S]$ denotes the assignment to the atoms in
$S$ as given by $M$. Then, every model $M$ of $\T_1$ must be, if we
    fix the atoms in $X$ to their values in $M$, a model of $\phi$ such that $M\models u$ if and only
if $\psi[M[X]]$ is unsatisfiable.

Starting from this result, we prove \PSPACE-hardness of inference $\T
\models \phi$ from a NAT $\T$.  The basic technique is to
introduce further variables as {\em parameters} $V$ into the formula $\Phi$
from
Proposition~\ref{lem:tcs},
which are kept fixed at the inner levels. At a new
outermost level to be added, the letter $u$ is used for evaluating the
formula at a certain level. We must in alternation minimize and
maximize the value of $u$.

Consider the case of a QBF $\Phi=\forall X\exists Y \psi[V]$, where $V$
are free variables in it, viewed as ``parameters''. We nest
$\T_1$ into the following theory $\T_2$: 
$$
\T_2= \{ X, Y, u: \T_1, u\lequiv \neg ab \}
$$
This amounts to the following circumscription: 
$$
\sigma(\T_2) = \exists ab.\circum(
\exists ab.\circum(\phi \land (u\!\lequiv\! ab);ab;Y,u)\land (u
\!\lequiv\!\neg ab);\,ab;\, X, Y, u).
$$
The outer circumscription minimizes $ab$ and thus maximizes $u$.
The formula
$\sigma(\T_2)$
is, by Proposition~\ref{prop:sigmaprime}, modulo the atoms $a_1$ and
$a_2$ equivalent to the formula
\begin{eqnarray*}
\sigma^{\star}(\T_2) &=& \circum(\sigma^{\star}(\T_1) \land (u \!\lequiv\! \neg a_2); a_2;
X, Y, u, a_1), \\ 
\sigma^{\star}(\T_1) &=& \circum(\phi \land (u\!\lequiv\!
    a_1);a_1;Y,u). 
\end{eqnarray*}
The following holds: 
\begin{proposition}
\label{lemma:qbf}
 $\T_2\models u$ if and only if for every truth assignment
$\nu$ to $V$, the QBF $\exists X\forall Y\neg \psi[\nu(V)]$ is true (i.e.,
$\Phi[\nu(V)]$ is false). 
\end{proposition}

\begin{proof}
$(\Leftarrow)$ Suppose $\T_2\not\models u$. Then, there exists a
model $M$ of $\T_2$ such that $M\models \neg u$. Since $M' \models
a_2$ holds for any model $M'$ of $\sigma^{\star}(\T_2)$ which extends
$M$ to $a_1, a_2$, we
conclude that every model $N$ of $\sigma^{\star}(\T_1) \land (u\lequiv
\neg a_2)$ such that $N[V]=M[V]$ satisfies $N\models a_2\land \neg u$ (otherwise, $N <_{a_2;X\cup Y\cup \{u,a_1\}} M$ would hold,
which contradicts that $M$ is a model of
$\sigma^{\star}(\T_2))$. Since $V\cup X$ is fixed in $\T_1$, it is
clear that
every assignment $\nu$ to $(V\cup X)$ which extends $M[V]$ can be
completed to a model $M_\nu$ of $\sigma^{\star}(\T_1)$. By minimality
of $M$, we have $M_\nu\models \neg u$, and thus $M\models \psi[\nu(V\cup X)]$. In other words,
$\forall X\exists Y \psi[\nu[V]]$ is true, which means that $\forall
X\exists Y \neg \psi[\nu(V)]$ is false for $\nu(V)=M[V]$.

$(\Rightarrow)$ Assume the assignment $\nu(V)$ is such that $\forall
X\exists Y \psi[\nu(V)]$ is true. Let $M$ be any model such
that $M[V]=\nu(V)$, $M\models \psi$, and $M \models a_2 \land \neg u
\land \neg a_1$. Then $M$ is a model of $\sigma^{\star}(\T_2)$. Indeed,
clearly $M$ is a model of $\sigma^{\star}(\T_1)$, since $M\models \phi
\land (u\lequiv a_1)$ and the minimized letter $a_1$ is false in
$M$. Furthermore, $M\models u \lequiv \neg a_2$. It remains to show
that there is no model $N$ of $\sigma^{\star}(\T_1)\land (u\lequiv
\neg a_2)$ such that $N <_{a_2;X\cup Y \{u,a_1\}} M$. Suppose such an
$N$ would exist. Then, $N\models u$, and we obtain that $\psi[N[V\cup
X]]$ is unsatisfiable. Since $N[V]=M[V]=\nu(V)$, this means that
$\forall X\exists Y \psi[\nu(V)]$ is false. This is a contradiction, and
thus $N$ can not exist. It follows that $M$ is a model of
$\sigma^{\star}(\T_2)$. Since $M\models \neg u$, the proposition is proved.
\end{proof}

A consequence of the preceding proposition is that deciding, given a NAT $\T_2$ of nesting
depth 1 and $\psi \in \Lang$, whether $\T_2 \models \psi$ is
$\PiP{3}$-hard.

We generalize this pattern to encode the evaluation of a QBF
\begin{equation}
\label{QBF}
\Phi = Q_n X_n Q_{n-1}X_{n-1}\cdots \forall X_2 \exists X_1 \psi,
\quad n\geq 1,
\end{equation}
where the quantifiers $Q_i$ alternate, into inference $\T \models
\psi$ from a NAT $\T$ as follows.

Let 
$\phi = \psi \lor u$, where $u$ is a fresh atom. Define inductively 
\begin{eqnarray*}
 \T_1 &=& \{ X_1,u : \phi, u \lequiv ab \}, \\
\T_{2k} &=& \{ X_1,\ldots,X_{2k}, u: \T_{2k-1}, u\lequiv \neg ab \},   \,\,\textrm{\ for all $2k \in \{2,\ldots, n\}$}, \\
\T_{2k+1} &=& \{ X_1,\ldots,X_{2k+1}, u: \T_{2k}, u\lequiv ab \},
                \quad \textrm{\ for all $2k+1 \in \{3,\ldots, n\}$},
\end{eqnarray*}
and let $\T_0 = \{~:
\phi\}$. Note that $\T_0$ is equivalent to $\phi$, and
that $nd(\T_i)=i-1$, for all $i\in \{1,\ldots,n\}$, while $nd(\T_0)=0$. 
We obtain the following. 
\begin{lemma}
\label{lem:nat}
For every $n\geq 1$ and possible truth assignment $\nu(X_n)$ to $X_n$,
$\T_{n-1}$ has some model extending $\nu(X_n)$, and 
\begin{itemize}
 \item if $n$ is odd, then $\T_{n-1} \models u$ if and only if $\Phi$ is
false, i.e., $\neg\Phi$ is true; 

\item if $n$ is even, then $\T_{n-1} \models \neg u$ if and only if
 $\Phi$ is true.
\end{itemize}
\end{lemma}

\begin{proof} The proof of this statement is by induction on $n\geq 1$.
For $n=1$, clearly $\T_0$ has for each truth assignment $\nu(X_1)$ some
model (just assign $u$ value true), and $\T_0\models u$ if and only if
$\Psi = \exists X_1\psi$ is false.  Suppose that the statement holds
for $n\geq 1$ and consider $n+1$.  Consider any truth assignment
$\nu=\nu(X_{n+1})$ to $X_{n+1}$, and let $\T_j^\nu$ be the NAT $\T_j$
for $\Phi^\nu = \Phi[\nu(X_{n+1})]$, $j\in \{ 0,\ldots,n\}$. Then, the
induction hypothesis implies that $\T_{n-1}^\nu$ has some model, which
can be extended to some model of $u\lequiv ab$ (resp., $u\lequiv \neg
ab$), and thus to all blocks in $\T_n^\nu$. Since the variables
$X_{n+1}$ are fixed in $\T_n$, also $\T_n$ must have a model which
extends $\nu(X_{n+1})$. Thus, the first part of the statement holds.

For the second part, assume first that $n+1$ is
odd. Then, $n$ is even, and by the induction hypothesis $\T^\nu_{n-1}\models
\neg u$ iff $\Phi^\nu$ is true. Since $u \lequiv\neg ab$ is a block of $\T_n$
and $X_n$ floats in $\T_n$ (while it is fixed in $\T_{n-1})$, it follows from
minimization of $ab$ that every model $M$ of $\T_n$ such that
$M[X_{n+1}]=\nu(X_{n+1})$ satisfies $u$ iff $\T^\nu_{n-1}\not\models \neg u$,
i.e., $\Phi^\nu$ is false.  Since the letters $X_{n+1}$ are fixed in $\T_n$, it
follows that $\T_n\models u$ iff $\neg\Phi^\nu$ is true for all truth
assignments $\nu(X_{n+1})$, which is
equivalent to $\Phi$ being false. Thus the statement holds in this case.

The case where $n+1$ is even is similar. By the induction hypothesis,
$\T^\nu_{n-1}\models u$ iff $\Phi^\nu$ is false. Since $u \lequiv ab$ is a
block of $\T_n$ and $X_n$ floats in $\T_n$ for minimizing $ab$, every
model $M$ of $\T_n$ such that $M[X_{n+1}]=\nu(X_{n+1})$ satisfies
$\neg u$ iff $\T^\nu_{n-1} \not\models u$, i.e., $\Phi^\nu$ is true. Since
$X_{n+1}$ is fixed in $\T_n$, it follows that $\T_n\models \neg u$ iff $\Phi^\nu$
is true for all truth assignments $\nu(X_{n+1})$, i.e., $\Phi$ is true. Thus, the statement holds
also in this case, which completes the induction step.
\end{proof}

We now turn to the problem of model checking. By our embedding of NATs
into nested circumscription, we obtain the following upper bound for
this problem. 

\begin{lemma}
\label{lem:modcheck}
Model checking for NATs, i.e., deciding whether a given interpretation
$M$ is a model of a given NAT $\T$, is in \PSPACE. If $nd(\T) \leq k$ for
constant $k\geq 0$, then it is in $\SigmaP{k+2}$.
\end{lemma}

\begin{proof}
By Proposition~\ref{prop:sigmaprime}, $M \models \T$ (thus
equivalently, $M\models\sigma(\T))$ if and only if there exists some
interpretation $M^*$ which extends $M[At\setminus Ab]$ to $A^*(\T)$ such that $M^*\models \sigma^{\star}(\T)$. By definition,
$\sigma^{\star}(\T) = \bigland_{i=1}^k A_i \land \bigland_{j=1}^m
\phi_j$ is a conjunction of circumscriptive atoms $A_i$ and ordinary
formulas $\phi_j\in \Lang((At\setminus Ab)\cup A^*(\T))$. Thus, we can decide $M \models \sigma(\T)$
by guessing a proper $M^*$ and check that $M^*\models A_i$ and
$M^*\models \phi_j$, for all $A_i$ and $\phi_j$.  We observe that
$nd(A_i) \leq k+1$ holds, since $nd(\sigma(\T))=nd(\sigma^{\star}(\T))
\leq nd(\T)+1$. Thus, by Theorem~\ref{theo:lcirc-mc}, each $M^*\models
A_i$ can be decided by a call to a $\PiP{k+1}$ oracle; deciding
$M^*\models \phi_j$ is polynomial, for every $\phi_j$.

Since $\sigma^{\star}(\T)$ and $A^*(\T)$ are constructible from $\T$
in polynomial time, it follows that deciding $M\models
\sigma^{\star}(\T)$, and thus $M\models \T$, is in $\SigmaP{k+2}$.
\end{proof}

The construction in Lemma~\ref{lem:nat} shows a polynomial-time
encoding of QBF evaluation into inference from a NAT. In turn,
Proposition~\ref{prop:sigmaprime} shows that a NAT can be polynomially
embedded into an $\Lcirc$ formula. The following theorem highlights
the consequences of such relations on complexity of inference with respect to  a
NAT.

\begin{theorem}
\label{theo:nat-inference}
Deciding, given a NAT $\T$ and a propositional formula $\phi$,
whether $\T\models \phi$ is \PSPACE-complete. If $nd(\T)\leq k$
for constant $k\geq 0$, then it is $\PiP{k+2}$-complete.
\end{theorem}

\begin{proof}
The hardness part follows from Lemma~\ref{lem:nat} above.  As for the
membership part, a model $M^*$ of $\sigma^{\star}(\T)$ such that
$M^*\not\models \phi$ (i.e., $M\not\models \phi$) can be guessed and
verified in \PSPACE\ (resp., with the help of a $\PiP{k+1}$-oracle in
polynomial time). Thus the problem is in co-NPSPACE = \PSPACE\ (resp.,
$\PiP{k+2}$).
\end{proof}

The complexity of NAT- satisfiability is now an easy corollary to the
previous results.

\begin{corollary}
\label{coroll:nat-sat}
Deciding whether a given NAT $\T$ is satisfiable is \PSPACE-complete. If
$nd(\T)\leq k$, for constant $k \geq 0$, then the problem is
$\SigmaP{k+2}$-complete.
\end{corollary}

The next theorem shows that the upper bounds on model checking for
NATs have matching lower bounds. For the general case, this is
expected from Theorem~\ref{theo:nat-inference}: if model checking
would be in PH, then also inference would be in PH. For the case of
bounded nestings, it turns out that compared to $\Lcirc$, the
minimization process of NATs has subtle effects on the complexity. In
particular, local abnormality letters are a source of complexity and
lift the problem, compared to similar $\Lcirc$ instances,  higher up
in PH. 
For example, in case $\T$ is a collection of blocks $B_i$ with nesting
depth zero, model checking for
$\T$ is $\SigmaP{2}$-complete, while for a corresponding conjunction of 
circumscriptive atoms $\circum(\phi_i;P_i;Z_i)$ where each $\phi_i$ is
an ordinary formula (having circumscriptive nesting depth
1), model checking is \CONP-complete.
%

\begin{theorem}
\label{theo:nat-mc}
Given a NAT $\T$ and an interpretation $M$, deciding whether
$M\models \T$ is \PSPACE-complete. If $nd(\T)\leq k$
for a constant $k\geq 0$, then the problem is
$\SigmaP{k+2}$-complete.%
\footnote{In the preliminary IJCAI '01 conference abstract of this
paper, incorrectly $\PiP{k+1}$-completeness of the problem was
reported. This result applies to a large natural subclass of theories
(which we had in mind), in particular, to theories which
allow polynomial model completion (see this section).} 
\end{theorem}

\begin{proof}
By Lemma~\ref{lem:modcheck}, it remains to show the hardness part.  To
this end, we use an extension of the encoding of a QBF in
Lemma~\ref{lem:nat}, and construct in polynomial time NATs $\T'_1,
\ldots, \T'_n$ and a model $M$ such that $M\models \T'_n$ iff the formula
$\Phi$ in (\ref{QBF}) for $n+1$ is true if $n$ is odd (resp., false if
$n$ is even).

Let the NATs $\T_1$,
\ldots, $\T_n$ be similar  as there, but with the following
                differences. Let $X_{n+1} = \{x_{n+1,1},$ $\ldots,$ $ x_{n+1,l}\}$.

\begin{itemize}
\item $\phi=\psi\lor u$
is replaced by $\phi'$, where 
$$
\phi' =\left\{
\begin{array}{ll}
(\psi \lor u \lor (X_n\land v)) \land ((X_n\land v)
\limplies u), & \mbox{if $n$ is odd,} \\
(\psi \lor u \lor (X_n\land v)), & \mbox{if $n$ is even.} 
\end{array}\right.
$$
Here $v$ is a new letter, which is described (i.e., floating) in $\T_n$ 
and fixed elsewhere. 

\item We add in $\T_n$ the formulas
\begin{eqnarray*}
\phi_{g} &:=& (X_n\land v) \rightarrow
\bigwedge_{j=1}^l \Big(ab_{n+1,j} \lequiv\neg ab'_{n+1,j}\Big)\quad \textrm{\ and\ }\\
\phi_{c} &:=& \neg (X_n\land v) \rightarrow \bigwedge_{j=1}^l \Big(x_{n+1,j} \lequiv
ab_{n+1,j}\Big),
\end{eqnarray*}
where $Ab_{n+1} = \{ab_{n+1,1},\ldots, ab_{n+1,l}\}$ and $Ab'_{n+1} =
\{ab'_{n+1,1},\ldots, ab'_{n+1,l}\}$ are fresh disjoint sets of
abnormality letters.

\item We describe (i.e., let float) the letters of $X_{n+1}$ in $\T_n$.

\end{itemize}
The resulting NATs, denoted $\T'_1,\ldots,\T'_n$, are thus as
follows. If $n=1$, then $\T'_1  = \{ X_1$, \ldots, $X_{n+1},$ $u, v: \phi', u\lequiv ab, 
\phi_g, \phi_c\}$; otherwise, 
\begin{eqnarray*}
\T'_1 &=& \{ X_1,u : \phi', \phi_{u,1} \}, \\
\T'_{j} &=& \{ X_1,\ldots,X_{j}, u: \T'_{j-1}, \phi_{u,j} \},
\qquad \textrm{for all $j \in \{2,\ldots, n-1\}$},\\
\T'_{n} &=& \{ X_1,\ldots,X_n,X_{n+1}, u, v: \T'_{n-1},\, \phi_{u,n}, \,
\phi_g,\, \phi_c\},
\end{eqnarray*}
where $\phi_{u,j} = u\lequiv ab$, if $j$ is odd, and $\phi_{u,j} =
u\lequiv \neg ab$, if $j$ is even. Note that $nd(\T'_n)=n-1$.

The intuition behind these modifications is as follows. Informally,
$X_n\land v$ will be true in a designated candidate model $M$, which enforces
that the value of $u$ is true if $n$ is odd (resp., false, if $n$ is
even by minimization of $ab$ in $\T'_1$). The candidate model $M$ can
only be eliminated by some other model which does not satisfy
$X_n\land v$, and thus must satisfy $\psi$.

Informally, $\phi_g$ serves for guessing a truth assignment $\nu$ to
the letters in $Ab_{n+1}$ for extending the designated model $M$ to a
witness $M^*$ for $M\models \T'_n$.  The assignment $\nu$ is
transfered by $\phi_c$ to $X_{n+1}$ when the minimality of $M^*$ is
checked; for that, it is assured that any possible smaller model $M'
<_{Ab;At\setminus Ab} M^*$ of the blocks in $\T'_n$ must falsify the
conjunction $X_n\land v$.

Define $M = \bigcup_{i=1}^n X_i \cup \{v,u\}$ if $n$ is odd and $M =
\bigcup_{i=1}^n X_i \cup \{ v\}$ if $n$ is even. Note that $M\models \phi'$. 

We claim that $M\models \T'_n$ iff the QBF $\Phi$ in (\ref{QBF}) for $n+1$  is false if $n$ is
odd (resp., true, if $n$ is even). Since $M$ and $\T'_n$ are
constructible in polynomial time, this will prove the result.

We use the following lemmas: 

\noindent{\bf Lemma A}. Let $M^*$ be any extension of $M$ such that
$M^*\models \phi_{u,n}$ and $M^*\models ab_{n+1,j} \lequiv
\neg ab'_{n+1,j}$, for all $j\in \{1,\ldots,l\}$. Then $M^*\models
\T'_{n-1}$ and $M^*\models ab$.

\noindent Proof: Note that $v$ and the letters in $X_n \cup X_{n+1}$
are fixed in $\T'_1,\ldots,\T'_{n-1}$. Thus, if any model $M'$ such
that $M'\models \T'_1$ coincides with $M$ on $X_n\cup \{v\}$, then it
follows $M'\models u$ if $n$ is odd (resp., $M'\models \neg u$, if $n$
is even). Next, all models $M'$ of $\T'_2$ which coincide with $M$ on
$X_n\cup\{v\}$ satisfy $M'\models u$ (resp., $M'\models \neg
u$). Continuing this argument, it follows that $M'[X_n\cup \{v\}] =
M[X_n\cup \{v\}]$ and $M'\models \T'_{n-1}$ implies that $M'\models u$
(resp., $M'\models \neg u$).
Hence, $M^*\models \T'_{n-1}$. Clearly $M^*\models ab$ holds.

\noindent{\bf Lemma B}. Let $M'$ be any model such that $M'\models
\phi_{u,n} \land \neg ab$ and $M'\not\models X_n\land v$. Then,
$M'\models \T'_{n-1}$ iff $\exists X_n \forall X_{n-1}\cdots
\exists X_1 \psi[M'[X_{n+1}]]$ is true if $n$ is odd, and $\forall
X_n \exists X_{n-1}\cdots \exists X_1 \psi[M'[X_{n+1}]]$ is false if
$n$ is even.

\noindent Proof: For any such $M'$, the problem $M'\models \T'_{n-1}$
is equivalent to $M'\models \T'_{n-1}$ (with the letters $X_{n+1}$ fixed
to their values in $M'$), since $X_n\land v$ in $\T'_1$ is false; the
new abnormality letters introduced above are irrelevant for
$M'\models \T_{n-1}$. Note that $M'\models \neg u$ if $n$ is odd
(resp., $M'\models u$ if $n$ is even). Lemma~\ref{lem:nat} implies
that $M'\models \T_{n-1}$ iff $Q_n X_n
Q_{n-1} X_{n-1}\cdots \exists X_1 \psi[M'[X_{n+1}]]$ is true if
$n$ is odd (resp., false if $n$ is even). This proves the lemma.

We now prove the claim. 

\noindent$(\Leftarrow)$ Suppose $M\not\models \T'_n$. Then, for each
extension $M^*$ of $M$ as in Lemma~A, there exists some model $M'
<_{Ab;At\setminus Ab} M^*$ of the blocks in $\T'_n$ such that
$M'\models \neg ab$, which implies that $M'\not\models X_n\land
v$. Hence, by Lemma~B, it follows that $Q_n X_n \forall X_{n-1}\cdots
\exists X_1 \psi[M'[X_{n+1}]]$ is true if $n$ is odd (resp., false if
$n$ is even). Since the different $M^*$ induce all possible truth
assignments to $X_{n+1}$, and $M'$ was arbitrary, it follows that
the formula $Q_{n+1} X_{n+1} Q_n X_n\cdots \exists X_1\psi$ is true if $n$
is odd (resp., false if $n$ is even).

\noindent$(\Rightarrow)$ Suppose that $M\models \T_n$. Hence, by
Proposition~\ref{prop:minimal-char} there exists a witness extension
$M^*$ of $M$ w.r.t.\ $\T_n$ which $(Ab;At\setminus Ab)$-minimally
satisfies the blocks in $\T_n$.  Thus, for each model $M'
<_{Ab;At\setminus Ab} M^*$ which coincides with $M$ on $Ab_{n+1}\cup
Ab'_{n+1}$ and such that $M'\models \neg ab$ and $M'\not\models
X_n\land v$, it follows that $M'\not\models\T'_{n-1}$. By Lemma~B, it
follows that $Q_n X_n \cdots \exists X_1 \psi[M'[X_{n+1}]]$ is false
if $n$ is odd (resp., true if $n$ is even), and thus $\exists X_{n+1}
\neg ( Q_n X_n\cdots \exists X_1 \psi[M'[X_{n+1}]])$ is true if $n$ is
odd (resp., false if $n$ is even).  Rewritten to prenex form, this is
means that the QBF in (\ref{QBF}) for $n+1$ is false if $n$ is odd
(resp., true if $n$ is even).
\end{proof}

We note that in the proof of the previous result, the fact that
abnormality letters are local to a NAT block plays an important role
for the complexity of model checking. The precise extension of these
letters is a priori unknown; an exponential search space may need to
be explored to find a suitable extension which satisfies the
propositional formulas in a block. By eliminating this source of
complexity, model checking becomes easier. This motivates the
following concept.

\begin{definition}
  We say that a block $B=\{ C: B_1,\ldots,B_m\}$ allows {\em polynomial model
   completion\/} if, given any model $M$ of $B$, a model $M^*$ is computable in
  polynomial time (as a function $f(M,B)$ of $M$ and $B$), such that $M^*$ is a
  witness extension of $M$ w.r.t.\ $B$ if $M\models B$.  A NAT $\T =
  B_1,\ldots,B_n$ allows {\em polynomial model completion}, if each block $B_i$
  allows polynomial model completion.
\end{definition}

Note that in general, assessing whether a block allows polynomial
model completion is a hard (intractable) problem. There are some
important cases, though, where this can be ensured. Namely, if the
abnormality letters $ab$ are used to minimize or maximize other
letters $p$, to which they are connected e.g.\ by equivalences
$ab\leftrightarrow p$ or inequivalences $ab \leftrightarrow \neg p$,
respectively (this will be further explored in
Section~\ref{sec:maximizing}). We obtain the following result.

\begin{theorem}
\label{theo:mc-pmc-block}
Let $B=\{ C: B_1,\ldots, B_m\}$ be any block that allows polynomial
model completion such that $nd(B)\leq k$, for constant $k\geq
0$. Then, model checking $M\models B$ is in $\DP_{k+1}$. Moreover, if
each $B_i\notin \Lang$ allows polynomial model completion, then
deciding $M\models B$ is in $\PiP{k+1}$.
\end{theorem}

\begin{proof}
Given $M$, by hypothesis and Proposition~\ref{prop:minimal-char}, we can complete it in polynomial time to a
model $M^*$ such that $M\models B$ iff $M^*$ is a $(Ab;C)$-minimal
model of $B_1,\ldots,B_m$. By Lemma~\ref{lem:modcheck},
each test $M\models B_i$ is in $\SigmaP{k+1}$. Furthermore, deciding
whether some model $M' <_{Ab;C} M^*$ exists such that $M'\models B_i$,
for $i=1,\ldots,m$ is in $\SigmaP{k+1}$; we can guess such an $M'$ and
for every $i\in\{1,\ldots,m\}$ a polynomial-size ``proof'' for
$M'\models B_i$ which can be checked with the help of a $\PiP{k}$
oracle in polynomial time. Thus, deciding $M\models B$ is reducible in
polynomial time to a conjunction of problems in $\SigmaP{k+1}$ and
$\PiP{k+1}$. Since these problems are in $\DP_{k+1}$ and this class is
closed under polynomial conjunctive reductions, it follows that
deciding $M\models B$ is in $\DP_{k+1}$. If each $B_i\notin
\Lang$ allows polynomial model completion, then by what we already
showed deciding $M\models B_i$ is in $\DP_{k}$ for every
$i=1,\ldots,m$. Thus, deciding whether no model $M' <_{Ab;C} M^*$
exists such that $M'\models B_i$ for all $i=1,\ldots,m$ is in
$\PiP{k+1}$, which means that $M\models B$ is reducible in polynomial
time to a conjunction of problems in $\PiP{k+1}$.  Since $\PiP{k+1}$
is closed under polynomial conjunctive reductions, it follows that
deciding $M\models B$ is in $\PiP{k+1}$.
\end{proof}

This membership result clearly generalizes from a single block to NATs
$\T = B_1,\ldots, B_n$ comprising multiple blocks, where each block
$B_i$ is as $B$ in the statement of
Theorem~\ref{theo:mc-pmc-block}. We remark that these upper bounds are
actually sharp, i.e., have matching lower bounds, but omit a proof of
this; for the case of nested polynomial model completion, a proof of
$\PiP{k+1}$-hardness is subsumed by the reduction in
Theorem~\ref{theo:nat-mc}, if we take $X_{n+1}$ to be empty (and thus
can eliminate the formulas $\phi_g$ and $\phi_c$ there).  

We note that Theorem~\ref{theo:mc-pmc-block} also shows that the
construction in the proof of Theorem~\ref{theo:nat-mc} uses
abnormality letters which are hard to complete in the right place
(thus revealing the source of complexity), namely in the outermost
block (and nowhere else). Indeed, moving them elsewhere would lead to
a decrease in complexity and the reduction would fail.

A final observation is that in the proof of the hardness part of
Theorem~\ref{theo:nat-inference}, the NATs $\T_j$ constructed allow
polynomial model completion. Thus, different from the case of model
checking, this property does not lower the complexity of inference
from NATs.

\section{Horn NATs}
\label{sec:horn-nat}

In this section, we consider a restricted class of NATs, which
generalizes Horn theories. Notice that Horn theories are an important
class of theories in knowledge representation, and the application of
the circumscription principle to Horn theories is underlying the
semantics of several logic programming languages, as well as
expressive database languages such as DATALOG$^{Circ}$
\cite{cado-palo-98}.

Recall that a clause is Horn, if it contain at most one positive
literal. 

\begin{definition}
We call a block $\{ C : B_1,\ldots, B_n\}$ {\em Horn}, if
each $B_i$ is a Horn CNF (i.e., a conjunction of Horn clauses) if $B_i
\in \Lang$, and recursively $B_i$ is Horn otherwise.  A NAT $\T$ is
{\em Horn}, if each of its blocks is Horn.
\end{definition}

\begin{example}\label{exa:horn}
{\rm NAT $\T$ in Example~\ref{exa:nat} is not Horn, because block $B$ contains
 non-Horn formula $b \wedge \neg ab \rightarrow f$. However, if we define
block $B'$ as:
\[ \{f:~~~ b \rightarrow f,~~~ c\rightarrow b,~~~ c \},
\]
then NAT $\T'$, defined as:
\[ \{f:~~~ f \rightarrow ab, ~~~B'\},
\]
is indeed Horn.  We can regard $B'$ as a ``simplified'' theory in which a
bird always flies.  \hfill$\Box$}
\end{example}

As for the complexity, it was shown in \cite{cado-lenz-94} that
deciding $\circum(\phi;P;\emptyset) \models \neg u$, where $\phi$ is a
propositional Horn CNF and $u$ is an atom, is $\CONP$-complete. As a
consequence, already for Horn NATs $\T$ without nesting (i.e.,
$nd(\T)=0$), inference is intractable.

We thus address the following two questions: Firstly, are there cases
under which (arbitrarily nested) Horn NATs are tractable, and
secondly, does nesting increase the complexity of Horn NATs?  In the
following subsection, we show that Horn NATs without fixed letters are
tractable, and that, fortunately, nesting does not increase the
complexity of Horn NATs. The latter result is not immediate and has
some implications for rewriting NATs, as will be discussed in
Section~\ref{sec:no-fixed}.

\subsection{Horn NATs without fixed letters} 
\label{sec:horn-nofixed}

In this subsection, we consider the fragment of Horn NATs in which no
fixed letters are allowed. That is, each letter $p$ except the special
abnormality letters must be described in any block. Note that in this
fragment minimization 
of letters $p$ is  still possible, via an
auxiliary atom $ab_p\in Ab$ and Horn axioms $p\rightarrow ab_p$,
$ab_p\rightarrow p$ which are included in the NAT.

We can view Horn NATs without fixed letters as a generalization of
(propositional) logic programs, which consist of Horn clauses
$a\leftarrow b_1,\ldots,b_n$, and whose semantics is given in terms of
the least (Herbrand) model, which amounts to parallel minimization of
all letters. By the above method, any such logic program $\Pi$ can be
easily transformed into a logically equivalent NAT; if $P$
is the set of letters, simply construct $\T_\Pi = \{ P: \Pi,\,
p\limplies ab_p,\, ab_p\limplies p,\ p\in P \}$, where $At=P\cup Ab$ and
$Ab=\{ab_p \mid p\in P\}$. However, NATs offer in addition nesting,
and furthermore some of the letters may float to minimize the
extension of other letters.

It is well-known that model checking and inference of literals from a
logic program is possible in polynomial time (cf.\
\cite{dant-etal-01}). It turns out that this generalizes to Horn NATs
without fixed letters, which can be regarded as a positive result. In
fact, as we shall show, any such NAT can be rewritten efficiently to a
logically equivalent Horn CNF.

In what follows, let us call any $(P;Z)$-minimal model of a NAT $\T$
such that $P=At\setminus Ab$ and $Z=\emptyset$ a {\em minimal model}
of $\T$.

\begin{theorem}
\label{theo:horn-nat-no-fixed}
Let $\T$ be a Horn NAT without fixed letters. Then, (i) $\T$ has the
least (i.e., a unique minimal) model $M(\T)$, and (ii) $\T$ is equivalent to some
Horn CNF $\phi(\T)$.  Furthermore, both $\phi(\T)$ and $M(\T)$ are
computable in polynomial time.
\end{theorem}

\begin{proof} 
Let, for any Horn CNF $\psi$ and interpretation $M$, be $\psi^M$ the Horn CNF
which results from $\psi$ after removing from it any clause which
contains some literal $(\neg) ab_j\in Ab$ such that $M\models (\neg)
ab_j$ and removing all literals $(\neg) ab_j$ such that $M\not\models
(\neg) ab_j$ from the remaining clauses. 

Let $\T$ be a single block $B=\{ Z: B_1,\ldots, B_n\}$, where
$Z=At\setminus Ab$. Define the Horn CNF $\phi(B)$ recursively by
\[
\phi(B) := \textstyle \bigwedge_{B_i\in \Lang} B_i^{M_0} \land \bigwedge_{B_i
\notin \Lang} \phi(B_i),
\]
where $M_0=M_0(B)$ is the least model of the Horn CNF 
\[
\psi(B) := \textstyle \bigwedge_{B_i\in \Lang} B_i \land \bigwedge_{B_i
\notin \Lang} \phi(B_i). 
\]
Furthermore, define  
\[
M(B) := M_0[At\setminus Ab]\ \  (=M_0[Z]). 
\]
Then, by induction on $nd(B)\geq 0$, we show that (i) $M(B)$ is
the least model of $B$, and (ii) $\phi(B)$ is logically equivalent to $B$.

(Basis) If $nd(B)=0$, then every $B_i$ is a Horn CNF, and both
$\phi(B)= \bigwedge_i B_i^{M_0}$ and $\psi(B)=\bigwedge_i B_i$ are
Horn CNFs. The block $B$ is equivalent to $\exists
Ab.\circum(\psi(B);Ab;Z)$, i.e., modulo $Ab$ to
$\circum(\psi(B);Ab;Z)$. Since $\psi(B)$ is a Horn CNF, it has the
$At;\emptyset$-least (i.e., a unique $(At;\emptyset)$-minimal) model
$M_0$. Notice that for every disjoint sets of atoms $P$ and $P'$ such
that $P\cup P' = At$ and any model $M$ of $\psi(B)$, it holds that
$M_0 \leq_{P;P'} M$.  Consequently, the projection $M(B) := M_0[Z]$ is
the unique minimal model of $B$. Thus item (i) holds for
$B$. Furthermore, if $M^*$ is a witness extension of any model $M$ of
$B$, then $M^*$ must coincide on $Ab$ with $M_0$, i.e.,
$M^*[Ab]=M_0[Ab]$. Thus, after fixing the value of each atom $ab_j \in
Ab$ as in $M_0$, the formula $\psi(B)$ describes all models of
$B$. That is, $\phi(B)= \bigwedge_i B_i^{M_0}$ is equivalent to $B$.
Thus item (ii) holds for $B$.

(Induction) Assume the statement holds for all $B$ with $nd(B) \leq
m$, and consider $m+1$. By the induction hypothesis, every $B_i$ in
$B$ is equivalent to $\phi(B_i)$. Thus, $B$ is equivalent to the block
$B' = \{ Z: B'_1,\ldots, B'_n\}$, where $B'_i=B_i$ if $B_i\in \Lang$
and $B'_i=\phi(B_i)$ if $B_i\notin \Lang$. Since $nd(B')=0$, by the
induction hypothesis $B'$ has the least model $M(B')$ and is
equivalent to $\phi(B')$.  Since $\psi(B)=\psi(B')$, we have
$M_0(B)=M_0(B')$ and $\phi(B)=\phi(B')$. Thus, the statement holds for
$B$, which concludes the induction step.

Let us now estimate the time needed for computing $M(\T)$ and
$\phi(\T)$, respectively.  For this purpose, let for any formula
$\alpha$, block $B$, NAT $\T$, etc denote $\|\alpha\|$, $\|B\|$,
$\|\T\|$ etc the representation size of the respective object.

Obviously, we can compute $\phi(B)$ bottom up. For the Horn CNFs
$\psi(B)$ and $\phi(B)$, we have $\|\psi(B)\| \leq \|B\|$ and
$\|\phi(B)\|\leq \|B\|$. Of the model $M_0$, we only need its
projection $M_0[V]$ to the set of atoms $V$ which occur in $B$; all
other atoms are irrelevant for computing $\phi(B)$. We can compute
$M_0[A]$ from $\psi(B)$ in $O(\|B\|)$ time; recall that the least
model of a Horn CNF $\alpha$ is computable in $O(\|\alpha\|)$ time,
cf.\ \cite{mino-88}. Furthermore, we can compute $\bigwedge_{B_i\in
\Lang} B^{M_0}_i$ from $M_0[A]$ in $O(\|B\|)$ time. Overall, it
follows that for $\T = B$, we can compute both $\phi(\T)$ and its
least model $M(\T)$ in time $O(\#b(\T)\|\T\|)$, where $\#b(\T)$ is the
number of (recursively occurring) blocks in $\T$, thus in polynomial
time.

By Proposition~\ref{prop:single-block}, we can replace a multiple
block NAT $\T = B_1,$ $\ldots,$ $B_n$ by the single block NAT $\T'= \{
At\setminus Ab : B_1,\ldots,B_n\}$, which is Horn and without fixed
letters, and obtain analogous results.
\end{proof}

Using sophisticated data structures, the (relevant parts of the)
models $M_0(B)$ in the proof of Theorem~\ref{theo:horn-nat-no-fixed}
can be computed incrementally, where each clause in $\phi(B)$ is fired
at most once. The data structures refine those used for computing the
least model of Horn CNF (see e.g.\ \cite{mino-88}).  Overall,
$\phi(\T)$ and $M(\T)$ are computable in $O(\|\T\|)$ time. We thus
have the following result:

\begin{theorem}[Flat Normal Form] 
\label{theo:flat-normal-form}
Every Horn NAT $\T$ without fixed letters can be rewritten to an
equivalent Horn NAT $\{ Z : \psi \}$ without fixed letters, where
$\psi \in \Lang$ is a Horn CNF, in $O(\|\T\|)$ time (i.e., in linear
time).
\end{theorem}

Thus, nesting in Horn NATs without fixed letters does not increase the
expressiveness, and can be efficiently eliminated. We remark that our
normal form result has a pendant in query languages based on fixpoint
logic (FPL), which is first-order predicate logic enriched with a
generalized quantifier for computing the least fixpoint of an
operator, defined in terms of satisfaction of a formula (see
\cite{gure-shel-86,imme-86} for details). It has been shown
\cite{gure-shel-86,imme-86} that over finite structures, nested use of
the fixpoint operator can be replaced by a single use of the fixpoint
operator. Our result, however, differs in several respects. FPL is an
extension to first-order logic, while strictly speaking, NATs are
second-order propositional theories. Furthermore, FPL has higher
expressiveness than the underlying logic, which is not the case for
Horn NATs without fixed letters. Finally, the complexity of rewriting
is not a concern in \cite{gure-shel-86,imme-86} which focus on the
existence of equivalent formulas without nestings, rather than on
efficient computation.

We note some easy corollaries of Theorem~\ref{theo:flat-normal-form}.

\begin{corollary}
\label{coroll:horn-nofixed-sat}
Deciding the satisfiability of a given Horn NAT $\T$ without fixed
letters is polynomial.
\end{corollary}

\begin{corollary}
\label{coroll:horn-nofixed-mcheck}
Model checking for a given Horn NAT $\T$ without fixed letters and
model $M$ is polynomial. 
\end{corollary}


The latter result will be sharpened in the next subsection.  For the
inference problem, we obtain the following result.

\begin{theorem}
\label{theo:horn-no-fixed}
Given a Horn NAT $\T$ without fixed letters and $\phi \in \Lang$,
deciding $\T \models \phi$ is $\CONP$-complete.  If $\phi$ is a CNF,
then the problem is polynomial.
\end{theorem}

\begin{proof} By Corollary~\ref{coroll:horn-nofixed-mcheck}, the problem is clearly in
$\CONP$. The $\CONP$-hardness part follows from $\CONP$-completeness
of checking the validity of a given formula $\phi\in \Lang$ (ask
whether $\{ Z: \psi \} \models \phi$, where $\psi$ is any tautology
and $Z$ contains all letters occurring in $\phi$).

We can reduce $\T\models \psi$ to $\phi(\T) \models \psi$ in
$O(\|\T\|)$ time, where $\phi(\T)$ is a Horn CNF. If $\psi =
\bigwedge_{i=1}^m\alpha_i$ is a CNF of clauses $\alpha_i$, the latter
can be checked in $O(m\|\phi(\T)\|+\|\psi\|)$ time, thus in
$O(m\|\T\|+\|\psi\|)$ time (check $\phi(\T)\models \alpha_i$, which
needs $O(\|\phi(\T)\|+\|\alpha_i\|)$ time, for all
$i\in\{1,\ldots.m\}$).
\end{proof}

\subsection{Horn NATs with fixed letters} 
\label{sec:horn-fixed}

The fragment of Horn NATs where fixed letters are allowed generalizes,
in a sense, the query language DATALOG$^{Circ}$ considered by Cadoli
and Palopoli \cite{cado-palo-98}.%
\footnote{Strictly, this applies to the propositional fragment of
DATALOG$^{Circ}$. The datalog setting of \cite{cado-palo-98} is
covered by the generalization of NATs to the first-order case
discussed in Section~\ref{sec:fo}.} In this language, circumscription
is applied to a conjunction of non-negative Horn clauses, which
describes an intensional database, viewing fixed predicates as
``free'' predicates for which any possible extension is considered,
while the other predicates are minimized or floating,
respectively. Thus, DATALOG$^{Circ}$ programs can be viewed as
unnested Horn NATs. 

As we have shown in the previous section, inference from a Horn NAT
without fixed letters is $\CONP$-complete, while model checking is
polynomial.  As we now show, the presence of fixed letters in Horn
nestings does not add complexity, i.e., reasoning stays
$\CONP$-complete and model checking remains polynomial.

These results build upon the fact that model checking for a Horn
circumscription $\circum(\phi;P;Z)$, which may have fixed letters, can
be polynomially reduced to model checking for a Horn circumscription
without fixed letters. Given an interpretation $M$, just check whether
$M$ is a model of $\circum(\phi \land \phi_{M,Q};P;Z\cup Q)$, where
$\phi_{M,Q}$ is a conjunction of literals that fixes the values of the
letters in $Q$ to the value as given in $M$. Clearly, the formula
$\phi \land \phi_{M,Q}$ is Horn.

Now the same method work recursively in a nested circumscription as
well; we end up with a Horn NAT that has no fixed letters. For such a
NAT, model checking is polynomial as we have shown in the previous
section. Overall, this means then that we have a polynomial time
procedure for model checking in the case of Horn NATs with fixed
letters.

More formally, we define the transformation $\alpha(M,B)$, where $M$
is any model and $B$ is either a formula from $\Lang$ or a block, as
follows:
$$
\alpha(M,B)=\left\{
\begin{array}{ll}
\phi, & \textrm{if } B=\phi \in \Lang; \\[1ex]
\{ Z\cup Q : \phi_{Q,M}, \alpha(M,B_1),\ldots,\alpha(M,B_m) \}, &
\textrm{if } B=\{Z: B_1,\ldots, B_m\} \\ 
 & \textrm{and } Q=At\setminus(Z\cup Ab) \textrm{ is}\\ 
 & \textrm{the set of fixed letters in } B, 
\end{array}
\right.
$$
where $\phi_{Q,M} = \bigland_{q\in Q \cap M} q \land \bigland_{q \in
Q\setminus M}\neg q$.
Furthermore, we define $$
\alpha(M,\T) = \bigland_{B\in \T} \alpha(M,B)
$$
for any interpretation $M$ and NAT $\T$. Observe that $\alpha(M,B)$
and $\alpha(M,\T)$ have no fixed letters. The following lemma states
that by the transformation $\alpha(M,\T)$, fixed letters can be
eliminated gracefully for the purpose of model checking.  

\begin{lemma}
\label{lem:elim-horn-fixed}
For any NAT $\T$ and interpretation $M$, we have that $M\models \T$ if and only
if $M\models \alpha(M,\T)$.
\end{lemma}

\begin{proof}
By definition of $M\models \T$, it remains to show that the statement
holds for any $\T$  which consists of a single block $B=\{
Z:B_1,\ldots,B_m\}$. This is accomplished by
induction on the nesting depth $n\geq 0$. 

(Basis) For $n=0$, we have $B_i\ = \phi_i \in \Lang$, for all
$i\in\{1\ldots,m\}$.  Suppose first that $M\models \alpha(M,B)$. By
Proposition~\ref{prop:minimal-char}, there is some witness extension
$M^*$ of $M$ w.r.t.\ $\alpha(M,B)$ which is an $(Ab;Z\cup Q)$ minimal
model of $B_1$, \ldots, $B_m$ and $\phi_{Q,M}$.  We claim that $M^*$
is a $(Ab;Z)$-minimal model of $B_1$,\ldots, $B_m$. Indeed, suppose
that some $M' <_{Ab;Z} M^*$ exists such that $M'\models B_1$, \ldots,
$M'\models B_m$. Since $M'\models \phi_{Q,M}$ must hold, it follows
that $M^*$ is not a $(Ab;Z\cup Q)$-minimal model of $B_1$, \ldots,
$B_m$ and $\phi_{Q,M}$. This is a contradiction. Thus, $M^*$ is a
$(Ab,Z)$-minimal model of $B_1,\ldots,B_m$. Hence, $M\models B$.

Conversely, assume that $M\models B$. Then, some witness extension
$M^*$ of $M$ w.r.t.\ $B$ is a $(Ab;Z)$-minimal model of
$B_1$,\ldots, $B_m$. By the definition, $M^* \models
\phi_{M,Q}$. Thus, $M^*\models \psi$ where $\psi = \phi_{M,Q}\land
B_1\land\cdots\land B_m$.  We claim that $M^*$ is a $(Ab;Z\cup
Q)$-minimal model of $\psi$. Towards a contradiction, assume that some
$M' <_{Ab;Z\cup Q} M^*$ exists such that $M' \models \psi$. Then, we
must have $M^*[Q]=M'[Q]$. Thus, $M'$ is a model of $B_1$,\ldots, $B_m$
such that $M' <_{Ab;Z} M^*$.  This means that $M\not\models B$, which
raises a contradiction. Therefore $M^*$ is an $(Ab;Z\cup Q)$-minimal
model of $\phi_{M,Q}$, $B_1$,\ldots, $B_m$. Consequently, $M\models
\alpha(M,B)$.  This proves the claim and concludes the case $n=0$.

(Induction) Suppose the statement holds for $n\geq 0$, and consider the case
$n+1$. Let $B = \{ Z: B_1,\ldots, B_m\}$. Then, $\alpha(M,B) = \{
Z\cup Q: \phi_{M,Q},\, \alpha(M,B_1),$ $\ldots, \alpha(M,B_m)\}$. By the
induction hypothesis, we have  that $M\models B_i$ iff
$M\models \alpha(M,B_i)$, for all $i\in \{1,\ldots, m\}$. Using similar
arguments as in the case $n=0$, we can see that
$M\models B$ holds precisely if $M\models \alpha(M,B)$ holds. 
\end{proof}

By combining Lemma~\ref{lem:elim-horn-fixed} and
Corollary~\ref{coroll:horn-nofixed-mcheck}, we thus obtain that model
checking for Horn NATs is polynomial.  A careful analysis of the
required computation effort reveals the following result.

\begin{theorem}
Model checking for Horn NATs, i.e., deciding whether $M\models \T$ for
a given interpretation $M$ and a Horn NAT $\T$, is possible in
$O(\|\T\|)$ time, i.e., in linear time in the input size.
\end{theorem}

\begin{proof}
A simple, yet not immediately linear time method is to check that $M
        \models B$ for each
block $B=\{Z: B_1,\ldots, B_m\}$ from $\T$ by exploiting
Lemma~\ref{lem:elim-horn-fixed} as follows:
\begin{enumerate}
\item recursively check that $M\models B_i$, for each $B_i\notin  \Lang$; 
\item compute the least model $M'_0$ of the Horn CNF $\psi'(B) =\phi_{Q,M}\land\bigwedge_{B_i \in \Lang} B_i$; 
\item check whether $M$ is a model of $\psi'(B)[M'_0[Ab]]$. 
\end{enumerate}
Note that this method is related to constructing the Horn
CNFs $\psi(B)$ and $\phi(B)$ for a Horn block $B$ without fixed
letters in the proof of Theorem~\ref{theo:horn-nat-no-fixed}.
Step~2 can be done in time $O(\max(|At|,\|\psi'(B)\|))$ and
Step~3 in time $O(\|\psi'(B)\|)$. These upper bounds, however, may
be reached and exceed $O(\|\{ Z: B_{i_1},\ldots,B_{i_l}
\}\|)$, where the $B_{i_j}$ are those blocks in $B$  which are not from
$\Lang$. If this happens recursively, the total time of the method
fails to be $O(\|B\|))$ as desired.

In Step 2, we can replace $\psi'(B)$ by $\psi''(B) =
\bigwedge_{B_i \in \Lang} B_i[M[Q]]$ and compute the least model $M''_0$ of $\psi''(B)$
on the letters occurring in it; this is feasible in $O(\sum_{B_i \in
\Lang}\|B_i\|)$ time.  In Step~3 then, we can replace
$\psi'(B)[M'_0[Ab]]$ by $\psi''(B)[M''_0[Ab]]$; 
checking whether
$M\models \psi''(B)[M''_0[Ab]]$ is feasible in $O(\sum_{B_i \in
\Lang}\|B_i\|)$ time.  Thus, the revised Steps 2 and 3 can be done in  $O(\|\{ Z: B_{i_1},\ldots,B_{i_l}
\}\|)$ time. This implies that checking $M\models B$ is feasible in
$O(\|B\|)$ time, from which the result follows. 
\end{proof}

Furthermore, we obtain from Lemma~\ref{lem:elim-horn-fixed},
Corollary~\ref{coroll:horn-nofixed-mcheck} and the intractability
result for Horn circumscription in \cite{cado-lenz-94} the following
result:

\begin{theorem}
Deciding, given a Horn NAT $\T$ and a propositional formula $\phi$,
whether $\T \models \phi$ is \CONP-complete. Hardness holds even if
$\T$ has nesting depth 0, and $\phi$ is a negative literal $\neg u$. 
\end{theorem}

This means that nesting is not a source of complexity for model
checking and inference from Horn NATs, which can be viewed as positive result. 

\section{Further Issues}
\label{sec:further}

In this section, we consider possible extensions of the results in the
previous sections to other representation scenarios. We first address
the class of $\Lcirc$ formulas and of NATs which do not have fixed
letters; as we have seen in the previous section, the presence of
fixed letters did not matter for the complexity of Horn NATs. We then turn
to a linguistic extension of NATs which has explicit maximization and
minimization of letters as primitives. While this extension does not
increase the expressiveness of NATs in general, it has some effects on
restricted NAT classes, and in particular on Horn NATs. Finally, we
briefly address the generalization of $\Lcirc$ and NATs to the predicate
logic setting.

\subsection{$\Lcirc$ formulas and NATs without fixed letters}
\label{sec:no-fixed}

In Sections~\ref{sec:lcirc-comp} and \ref{sec:nat-comp}, we have
considered $\Lcirc$ formulas and NATs in a general setting which
allows for fixed letters in circumscriptions, and we have seen in the
previous section that the presence of fixed letters does not matter for the
complexity of Horn NATs.

As shown below, fixed letters can be removed from $\Lcirc$ and NAT
theories, respectively, by simple techniques. By exploiting them, the
hardness results of Sections~\ref{sec:lcirc-comp} and
\ref{sec:nat-comp} can be sharpened to theories without fixed letters.

\subsubsection{Eliminating fixed letters from a $\Lcirc$ formula}
\label{sec:elim-fixed-lcirc}

De Kleer and Konolige have shown \cite{dekl-kono-89} a simple
technique for removing the fixed letters from an ordinary
circumscription. The same technique can be applied for formulas from
$\Lcirc$ as well.  More precisely, let $\phi = \circum(\psi;P;Z)$ be a
circumscriptive atom. Then,
\begin{enumerate}
\item For each letter $q\notin P\cup Z$, introduce a fresh letter
$q'$, and add both $q$, $q'$ to
$P$;
\item add a conjunct $q \lequiv
\neg q'$ to $\psi$.
\end{enumerate}
Let $\phi' = \circum(\psi';P';Z)$ be the
resulting circumscriptive atom.  Then, the following holds. 

\begin{proposition}
Modulo the set of all auxiliary
letters $q'$, the formulas $\phi$ and $\phi'$ are logically equivalent.
\end{proposition}

Using this equivalence, we can eliminate all fixed letters from a
formula $\alpha \in \Lcirc$, by replacing each circumscriptive atom
$\phi$ in $\alpha$ with $\phi'$, where the fresh atoms $q'$ are
made minimized inside $\phi'$ and outside $\phi$. Note that the
resulting formula $\alpha'$ has size polynomial in the size of
$\alpha$.

\subsubsection{Eliminating fixed letters from a NAT}
\label{sec:elim-fixed-nat}

Every fixed letter $q$ can be removed from a NAT $\T$ similarly as
from a formula $\phi \in \Lcirc$. However, we must take into account
that a fixed letter $q$ may not be simply declared as a minimized
letter in the rewriting, since there is the special set $Ab$ of
minimized letter which has restricted uses. We surpass this as
follows:
\begin{enumerate}

\item Introduce, for each fixed letter $q$, two special abnormality letters $ab_q$ and
$ab'_q$ in $Ab$;

\item add the formula $(q \lequiv ab_q)\land (ab_q \lequiv
\neg ab'_q)$ as a new block in each block $B$ occurring in $\T$ where $q$
is fixed;

\item 
declare $q$ as described (i.e., floating) in each block occurring in $\T$.
\end{enumerate}

Let $\T'$ be the resulting NAT (which has an extended set of
abnormality letters, $Ab\cup Ab'$). Then, we have:

\begin{proposition}
Modulo the set $Ab'$ of auxiliary letters, the NATs $\T$ and $\T'$ are
logically equivalent, i.e., have the same sets of models. 
\end{proposition}

Note that the rewriting adds $O(|At|)$ symbols in each block,
and is feasible in $O(|At|\!\cdot\!\#b(\T))$ time, where $\#b(\T)$ is the
number of (recursively occurring) blocks in $\T$. Furthermore, observe
that the method uses non-Horn clauses.  This is not accidently; from
the tractability result for inference of a CNF from a Horn NAT without
fixed letters (Theorem~\ref{theo:horn-no-fixed}) and the
intractability of inference of a literal from a Horn
circumscription \cite{cado-lenz-94}, we can infer that there is no
simple polynomial-time rewriting method which uses only Horn clauses,
unless $\Pol=\NP$. This is also possible if we allow $\T'$ to be any
Horn NAT without fixed letters (not necessarily equivalent) and the
query to be replaced by any CNF $\phi'$, such that $\T \models \phi$
is equivalent to $\T' \models \phi'$ for the query $\phi$ at hand.

\subsection{Maximizing and minimizing predicates}
\label{sec:maximizing}

In his seminal paper \cite{lifs-95}, Lifschitz discussed two explicit
constructs $\min p$ and $\max p$ for defining a minimal and a maximal
extension of a letter $p$ in a NAT, respectively. These constructs
are easily implemented by using designated abnormality letters.

\begin{definition} An {\em extended block} is any expression
\begin{equation}
\label{x-nat}
\{ C ;\, \min C^-;\, \max C^+ : B_1,\ldots, B_m \},
\end{equation}
where $C$, $C^-$, and $C^+$ are disjoint sets of atoms from
$At\setminus Ab$; if empty, the respective component is omitted.
Intuitively, the letters in $C$ are defined as usual while for those
in $C^-$ (resp., $C^+$), a minimal (resp., maximal) extension is
preferred. An {\em extended NAT} is a collection $\T=B_1,\ldots,B_n$
of extended blocks.
\end{definition}

\begin{example}\label{exa:diagnosis}
{\rm Let us consider model-based diagnosis at a superficial level.  In
Reiter's approach \cite{reit-87}, a diagnosis problem consists of a
system description $SD$, a set of observations $\mathit{OBS}$ (which
are facts), and a set of components $\mathit{COMP} = \{ c_1,$
$\ldots,$ $c_m\}$ in the system. $SD$ is a set of axioms which
describe the structure and the functioning of the system, using
designated atoms $ok_i$ which informally expresses that component
$c_i$ works properly. A {\em diagnosis} is a minimal set
$\Delta\subseteq C$ such that $SD \cup \{\mathit{OBS}\} \cup \{ \neg ok_i
\mid c_i\in \Delta \} \cup \{ ok_j \mid c_j\in
\mathit{COMP}\setminus\Delta\}$ is satisfiable. That is, $\Delta$
assumes as little malfunctionings as needed to explain the
observations (equivalently, as many components as possible are assumed
to work properly). 

Assuming a modular system design, each component $c_i$ may be
represented as a block $B_i = \{ ok_i, V_i: \ldots\}$, where inputs
are passed to $B_i$ via variables that are fixed, and outputs from $c_i$ are modeled
by variables $V_i$ which are described, together with a variable
$ok_i$ which indicates whether $B_i$ works properly.
The components may be linked by some axioms $\phi_1,$
$\ldots,$ $\phi_n$, such that $B= \{ V;\, \max ok_1, \ldots, ok_m:
\phi_1,$ $\ldots$, $\phi_n, B_1,$ $\ldots, B_m\}$ represents the
system. Then, the models of $\T = B, \{~: \mathit{OBS} \,\}$ correspond to the
diagnoses of the system. If a block $B_i$ is hierarchically composed,
further nesting of blocks may be used in the modeling.

As an example, consider the following very simplified model of a Web server for
electronic commerce, composed of two modules:
\begin{enumerate}
\item an application server, with features for the client interface and the
  interaction with the database system,
\item a database system storing data on customers, orders, etc.\ with a query
  that must be executed on it for each interaction with the client.
\end{enumerate}
The modules can be, respectively, modeled by means of the following blocks:
\begin{enumerate}
\item $B_1 = \{ ok_1, V_1:~~ ci \land db \land ok_1 \rightarrow V_1\}$, where
  $ci$ and $db$ mean, respectively, that the interaction with the client and the
  database system have been performed;
\item $B_2 = \{ ok_2, V_2:~~ V_1 \land q  \land ok_2 \rightarrow
  V_2, ~~ V_1 \wedge \neg q \rightarrow \neg ok_2
\}$, where
  $q$ means that the  query has been executed.
\end{enumerate}
Description of the entire system can be made by means of the following block:
\[
B = \{ V, V_1, V_2;\, \max ok_1, ok_2:~~ V_1 \land V_2 \rightarrow V,
~B_1, ~B_2\},
\]
where $V$ is a new symbol.  The above description can be used, for example,
during the test phase of the Web server, in which interaction with one client
is simulated. During such a phase, an administrator checks whether the query
and the interactions between modules have been performed. Now, assume that the
administrator determines that interactions with the client and the database
system have been performed, but the query has not been executed, i.e., the set
of his observations is $\mathit{OBS} = \{ ci, db, \neg q \}$. It is easy to determine that the
diagnoses of the system correspond to either $\{\neg ok_1\}$, or $\{\neg
ok_2\}$, i.e., one subsystem is malfunctioning, but not both.  \hfill$\Box$ }
\end{example}

Other examples for the use of maximization can be found in \cite{lifs-95}.

Formally, the semantics of an extended block $B$ as in (\ref{x-nat})
can be defined by a transformation $(\cdot)^\circ$ to the ordinary
block
\begin{equation}
\label{x-nat-sem}
B^\circ = \{ C \cup C^- \cup C^+ : \phi_{C^-}, \phi_{C^+}, B_1,\ldots, B_m \},
\end{equation}
where $\phi_{C^-}= \bigland_{p\in C^-}(p \rightarrow ab_p)$ and
$\phi_{C^+} = \bigland_{p\in C^+}(\neg ab_p \rightarrow p)$, and 
each $ab_p$ is an abnormality letter not used in any $B_i$ which is a
formula from $\Lang$.  For any extended NAT $\T=B_1,\ldots,B_n$, we then define
$\T^\circ = B_1^\circ$, \ldots, $B_n^\circ$. 

Thus, the constructs $\min$ and $\max$ do not increase the
expressiveness of NATs in general. However, we have a different
picture in restricted cases. In particular, maximization of letters
increases the expressiveness of Horn NATs.  As follows from the next
theorem, extended Horn NATs climb the levels of PH closely behind
general NATs as the nesting depth increases (at one level distance for
inference and satisfiability, and at two levels for model checking),
and they are \PSPACE-complete for unbounded nesting depth. Note that
maximization in Horn NATs is useful. In the diagnosis application of
Example~\ref{exa:diagnosis}, the axioms describing the system might be
Horn; for example, the clauses $in_1\land in_2 \land ok_g \limplies
out$, $out\land ok_g \limplies in_1$, and $out\land ok_g\limplies in_2$
may describe a logical and-gate $g$ whose output is true, if working
properly, exactly if both inputs are true. Note that in the particular
diagnostic Web-Server scenario, all formulas in $\T$ are Horn
clauses except $V_1\land \neg q \rightarrow \neg ok_2$, which can be easily rewritten to a
Horn clause. 

We need some auxiliary results, which are of interest in their own
right.  In what follows, we denote for any block $B$ by $SBl(B)$ the
set containing $B$ and all blocks $B'$ that recursively occur in $B$,
and for any NAT $\T=B_1,\ldots,B_n$, we define $SBl(\T)
=\bigcup_{i=1}^n SBl(B_i)$.

\begin{proposition}
\label{prop:mc-rpmc-block}
Let $B$ be any block such that for every $B'\in SBl(B)$, (i) $B'$ allows polynomial
model completion, and (ii) model checking $M\models B'$ is polynomial if
$nd(B')=0$. Then model checking $M\models B$ is in $\PiP{k}$.
\end{proposition}

\begin{proof}
The proof is similar to the proof of the $\PiP{k+1}$-membership part in
Theorem~\ref{theo:mc-pmc-block} for the case of polynomial model completion,
but exploits that in the base case ($nd(B)=0$), model checking is
polynomial rather than in $\CONP=\PiP{1}$. 
\end{proof}

\begin{proposition}
\label{prop:xhorn-pmc}
Let $\T$ be any extended Horn NAT. Then, every block $B\in SBl(\T)$ allows polynomial model completion.
\end{proposition}

\begin{proof}
Let $B = \{ C;\, \min C^-;\, \max C^+: B_1,\ldots, B_m\}$. Without loss of generality, we assume that $C^-$ is empty, i.e., the
$\min$-part
is missing: since the formula $\phi_{C^-}$ is Horn, we may add it in
$B$ while keeping the Horn property and move $C^-$ to the ordinary
defined letters $C$.  Suppose that $B_1, \ldots,B_l$ ($l\leq m$) are
all blocks $B_j$ that are formulas (i.e., $B_j\in\Lang$), and let
$\psi$ be their conjunction. Let $M$ be the model to be
completed. Define
$$
\psi' = \psi \land \bigland_{p \in (At\setminus Ab) \cap M} p \land
\bigland_{p \in C^+ \cap M} ab_p  \land \bigland_{p\in At\setminus (Ab\cup
M)} \neg p \land \bigland_{p\in C^+\setminus M} \neg ab_p\,. 
$$
Note that $\psi'$ is Horn, and thus, if satisfiable, it has the unique
least model $M'$, which obviously coincides with $M$ on the atoms in
$(At\setminus Ab) \cup \{ab_p\mid p\in C^+\}$ and is computable in
polynomial time.

Consider the transformed block $B^\circ = \{ C\cup C^+: \phi_{C^+},
B_1,\ldots, B_m \}$. We claim that $M\models B$ if and only if $M'$ is
a $(Ab;C\cup C^+)$-minimal model of $\phi_{C^+}$, $B_1$,\ldots, $B_m$.
By Proposition~\ref{prop:minimal-char}, the if-direction is
immediate. For the only-if direction, suppose that $M\models B$. Then,
by Proposition~\ref{prop:minimal-char}, there exists a witness
extension $M^*$ of $M$ which is a $(Ab;C\cup
C^+)$-minimal model of $\phi_{C^+}$, $B_1$,\ldots, $B_m$. Since $M$
and $M'$ coincide on $At\setminus Ab$ and each atom $ab_p$, $p\in
C^+$ occurs only in $\phi_{C^+}$, the minimality of $M^*$ implies
that $M^*$ and $M'$ coincide on $(At\setminus Ab) \cup \{ab_p \mid
p\in C^+\}$ and thus $M^*\models \psi'$. Since $M'$ is the least model
of $\psi'$, it follows $M'\leq_{Ab;C\cup C^+} M^*$. By construction,
$M'\models \phi_{C+}$, and $M^*\models B_i$ implies that $M'\models
B_i$, for each block $B_j$ where $j\in \{l+1,\ldots,m\}$. Thus, from the
minimality of $M^*$, we conclude that $M'=M^*$. This proves the claim.
\end{proof}

\begin{theorem}
\label{theo:xhorn-mc+inf}
For extended Horn
theories $\T$, (i) model checking $M\models \T$, (ii) inference $\T\models
\phi$, and (iii) deciding satisfiability are \PSPACE-complete. Furthermore,
(i) is polynomial if $k=0$ and $\PiP{k}$-complete if $k\geq 1$, (ii) is $\PiP{k+1}$-complete, and (iii) 
is $\SigmaP{k+1}$-complete, if $nd(\T)\leq k$ for a constant $k\geq 0$.
\end{theorem}

\begin{proof}
For the membership parts, by Theorems~\ref{theo:nat-inference},
\ref{theo:nat-mc} and Corollary~\ref{coroll:nat-sat}, it remains to
show the statement for bounded nesting depth. From
Propositions~\ref{prop:xhorn-pmc} and \ref{prop:mc-rpmc-block}, this
is easily seen to hold, provided that model checking $M\models B$ for
any extended Horn block $B = \{ C;\, \max C^+: B_1,\ldots, B_m\}$ such
that $nd(B)=0$ is polynomial.

To prove the latter, let $M'$ be the least model of the Horn CNF
$\psi'$ constructed from $B$ in the proof of
Proposition~\ref{prop:xhorn-pmc}. As shown there, $M\models B$ iff
$M'$ is a $(Ab;C\cup C^+)$-minimal model of $\phi_{C^+}$,
$B_1$,\ldots, $B_m$. We can check $M'\models \phi_{C^+}$ and
$M'\models B_i$ for all $B_i$ easily in polynomial time. Furthermore,
we can check $(Ab;C\cup C^+)$-minimality of $M'$ by testing whether
each of the following Horn CNFs $\psi_\ell$ is unsatisfiable. Let $F^+
= M\cap (At \setminus (Ab \cup C))$ and $F^- = At \setminus (M \cup C
\cup C^+ \cup \{ ab_p \mid p \in C^+\})$. For each literal $\ell \in
(C^+\setminus M) \cup \{ \neg ab \mid ab \in  M \cap (Ab \setminus\{ ab_p\mid p\in C^+\})\}$, define
$$
\psi_\ell \,=\, \ell \land \bigwedge_{p\in F^+} p \land \bigwedge_{p\in F^-}\!\!\neg p \land \bigwedge_{i=1}^m B_i;
$$
that is, we fix the ``interesting'' letters which are not defined in
$B$ to their values in $M'$, fix each letter from $C^+$ which is true
in $M'$, and fix each ``regular'' abnormality letter (not introduced
for a letter in $C^+$) which is false in $M'$; furthermore, $\ell$
serves to increase one letter in $C^+$ (resp.\ decrease one regular
abnormality letter) compared to $M'$. Thus, no model $M''$ exists such
that $M''<_{Ab;C\cup C^+}M'$ iff each $\psi_\ell$ is unsatisfiable,
which can be checked in polynomial time. In summary, testing whether
$M'$ is an $(Ab;C\cup C^+)$-minimal model of $\phi_{C^+}$,
$B_1$,\ldots, $B_m$, and thus whether $M\models B$, is possible in
polynomial time. This concludes the proof of the membership parts.

The hardness proofs for (i) and (ii) are obtained by slight
modifications of the reductions in the proofs of
Theorems~\ref{theo:nat-mc} and \ref{theo:nat-inference} (i.e.,
Lemma~\ref{lem:nat}). The hardness proof for (iii) follows from the
hardness proof of (i), since the formula $\phi$ in the reduction is a
single literal and $\T\models \phi$ iff the NAT $\T, \{~: \neg \phi \}$
is unsatisfiable.

For (ii), the modifications to the NATs $\T_1,\ldots,\T_n$ in the
proof of Theorem~\ref{theo:nat-inference} are as follows: 
\begin{enumerate}
\item Drop in each $\T_{2k+1}$ (resp., $\T_{2k}$) the formula
$u\leftrightarrow ab$, (resp., $u\leftrightarrow \neg ab$), and
declare $u$ minimized (resp., maximized). 

\item Introduce for each letter $p \in At \setminus (Ab \cup \{ u\})$ ($=:A$)
a fresh letter $p'$; intuitively, $p'$ serves for emulating the
negation of $p$. This is accomplished by adding in $\T_1$ an extended Horn block
$
B_{\not\equiv} = \Big\{ \max A, A': \bigwedge_{p\in A} (\neg p \lor\neg p')\Big\}.
$
Informally, the parallel maximization of $p$ and $p'$ generates two
models; one has $p$ true and $p'$ false, and the other has vice versa
$p'$ true and $p$ false. In this way, $p'$ is defined as the
complement of the $p$. 

\item We replace in $\T_{1}$ the formula $\phi=\psi \lor u$ by the Horn CNF
$
\hat{\phi}' = \bigwedge_{j=1}^l (\gamma'_j \lor u),
$ 
where  w.l.o.g.\
$\psi = \bigwedge_{j=1}^l\gamma_j$ is conjunction of clauses and
$\gamma'_j$ results from $\gamma_j$ by replacing each positive
literal $x$ by the negative primed literal $\neg x'$.

\item We let $p'$ be described in the same
NATs $\T'_j$ where $p$ is described, for each $p\in A$.
\end{enumerate}
The resulting NATs, denoted $\hat{T}_1$, \ldots, $\hat{T}_n$, are thus as follows:
\begin{eqnarray*}
 \hat{T}_1 &=& \{ X_1, X'_1;\, \min u : \hat{\phi},\, B_{\not\equiv} \}, \\
 \hat{T}_{2k} &=& \{ X_1,X'_1, \ldots,X_{2k}, X'_{2k}, \max u:
 \hat{T}_{2k-1} \}, \quad~~~~ \textrm{ for all $2k \in \{2,\ldots, n\}$, }\\
 \hat{T}_{2k+1} &=& \{ X_1,X'_1,\ldots,X_{2k+1},X'_{2k+1};\, \min u:
 \hat{T}_{2k}\}, \quad \textrm{ for all $2k+1 \in \{3,\ldots, n\}$. }
\end{eqnarray*}
Observe that $nd(\hat{T}_n) = n$. It is easily seen that modulo the
new letters, $\hat{T}_j$ and $\T_j$ have the same models, for
$j=\{1,\ldots,n\}$. Thus, the hardness result for (ii) follows.

For (ii), the modifications to the NATs $\T'_1,\ldots,\T'_n$ in the
proof of Theorem~\ref{theo:nat-mc} are similar to those in (i), but
with the following differences:
\begin{itemize}
\item We perform the reduction with empty $X_{n+1}$, i.e., we suppress
the leading quantifier $Q_n X_{n+1}$; the formulas $\phi_g$ and
$\phi_c$ are removed from $\T'_n$ (they are tautologies).

\item In step 3, instead of $\phi=\psi\lor u$ we replace in $\T'_{1}$ the formula $\psi \lor u \lor (X_n\land
v)$ by the Horn CNF
$
\hat{\phi}' = \bigwedge_{j=1}^l\bigwedge_{x\in X_n \cup \{ v\}} (\gamma'_j
\lor \neg x' \lor u).
$ 
\end{itemize}
The resulting NATs, denoted $\hat{T}_j$, are for odd $n > 1$ thus
as follows (for even $n$, they are analogous): 
\begin{eqnarray*}
 \hat{T}'_1 &=& \{ X_1, X'_1;\, \min u : \hat{\phi}',\, (X_n\land v)
 \rightarrow u,\, B_{\not\equiv} \}, \\ \hat{T}'_{2k} &=& \{ X_1,X'_1,
 \ldots,X_{2k}, X'_{2k};\, \max u: \hat{T}'_{2k-1} \}, \quad~~~
 \textrm{ for all $2k \in \{2,\ldots, n-1\}$, }\\ \hat{T}'_{2k+1} &=&
 \{ X_1,X'_1,\ldots,X_{2k+1},X'_{2k+1};\, \min u: \hat{T}'_{2k} \},
 \quad \textrm{ for all $2k+1 \in \{3,\ldots, n-1\}$,}\\[1ex]
 \hat{T}'_n &=& \{ X_1,X'_1,\ldots,X_n,X'_n, v,v';\, \min u:
 \hat{T}'_{n-1} \}.
\end{eqnarray*}
Notice that $nd(\hat{T}'_n) = n$. Modulo the new letters, $\hat{T}'_j$
and $\T'_j$ have the same models, for $j\in\{1\ldots,n\}$. Thus, the hardness result for (i)
follows.
\end{proof}

Note that model checking for extended Horn NATs resides in PH two
levels below arbitrary NATs of the same nesting depth. The proof
reveals that this can be ascribed to the benign properties that both
model completion and polynomial-time model checking for an extended
Horn circumscription (where maximization of letters besides
minimization is allowed) are polynomial. Each of these tasks is a
source of complexity, i.e., intractable for arbitrary NATs. In
particular, for a collection of unnested extended Horn blocks, model
checking is polynomial and inference is $\CONP$-complete, which means
that the latter can be polynomially transformed to a SAT solver.
Likewise, for nesting depth 1, inference is $\PiP{2}$-complete, and
thus polynomially reducible to inference from an ordinary (non-Horn)
circumscription, as well as to engines for knowledge representation
and reasoning which are capable of solving $\PiP{2}$-complete
problems, such as {\sc dlv} \cite{eite-etal-99h,dlv-web}.

We finally remark that using maximization, fixed letters can be easily
eliminated from extended Horn NATs similarly as from general NATs.
(Namely, introduce in each block $B$ for every fixed letter $q$ a fresh letter
$q_B$, and add the clause $\neg q\lor\neg q_B$ in $B$ and declare $q_B$ and
$q$ maximized; in all other blocks, let both $q$ and $q_B$ float.)
Thus, the complexity results for extended Horn NATs from above can be
strengthened to theories without fixed letters.

\subsection{First-order case}
\label{sec:fo}

In this paper, we have considered so far nested circumscription and
NATs in a propositional language. There is no difficulty in extending
the language $\Lcirc$ to the case of first-order predicate logic,
along the definition of second-order parallel circumscription of
predicates \cite{lifs-85a,lifs-94}; the formulation of NATs in 
\cite{lifs-95} is actually for predicate logic.

As shown by Schlipf \cite{schl-87}, and further elaborated on in
\cite{cado-etal-92}, circumscription is capable of expressing problems
at the $\Sigma^1_2$ and $\Pi^1_2$ level of the prenex hierarchy of
second-order logic, and thus highly expressive far beyond the
computable. Thus, also nested circumscription and NATs are highly
undecidable in the general first-order setting. However, decidable
fragments can be obtained by imposing suitable restrictions. 

An important such fragment is given if the
theories include a {\em domain closure axiom}
\begin{equation}
\label{DC}
\mathrm{(DCA)}\quad \forall x.(x=c_1\lor x=c_2\lor\cdots\lor x=c_n),
\end{equation}
where $c_1,\ldots,c_n$ are the (finitely many) constant symbols available,
and the {\em unique names axioms}
\begin{equation}
\label{UNA}
\mathrm{(UNA)}\quad c_i\neq c_j, \qquad \mbox{for all $i\in \{1,\ldots,n\}$ and $j\in \{
i+1,\ldots, n\}$.} 
\end{equation}
Such a setting is quite popular in KR and, in the absence of function
symbols, in deductive databases, where it is also known as the
``datalog'' setting. It is essentially propositional, where in the
datalog setting models correspond to Herbrand models over the given
alphabet. The setting allows for a more compact representation, which
on the other hand may lead to an exponential complexity increase. This
is reflected in the complexity of $\Lcirc$ and NATs in this setting.

\begin{theorem}
Inference and satisfiability of a first-order 
$\Lcirc$  formula (resp., NAT $\T$) under DCA and UNA is {\rm
EXPSPACE}-complete.
\end{theorem}

The upper bounds are straightforward by reducing a $\Lcirc$ formula
(resp., NAT) to its equivalent ground instance, which is
propositional and constructible in exponential time; functions
$f(x_1,\ldots,$ $x_n)$ can be eliminated, as well-known, with polynomial
overhead by introducing fresh predicates $F(x_1,$ $\ldots$, $x_n,y)$ and axioms
$\forall x_1 \cdots x_n!\exists y.F(x_1,\ldots,x_n,y)$ such that
$\lambda y F(x_1,\ldots,x_n,y)$ amounts to $\lambda
y(y=f(x_1,\ldots,x_n))$. The lower bounds for these results are
obtained by a straightforward generalization of the QBF encoding in
Lemma~\ref{lem:nat} to encodings of sentences $Q_n P_n Q_{n-1}
P_{n-1}$ $\cdots$ $\forall P_2\exists P_1\psi$ of second-order logic, where
each $P_i$ is a list of predicate variables of given arities and
$\psi$ is function-free first-order. For bounded nesting
depth, the complexities parallel the respective levels of PH at its
exponential analogue, the Weak EXP Hierarchy ({\rm EXP}, {\rm NEXP},
{\rm NEXP}$^{\NP}$, {\rm NEXP}$^{\SigmaP{2}}$,\ldots). For example,
inference $\phi\models \psi$ of $\Lcirc$ sentences $\phi$ and $\psi$
is co-NEXP$^{\SigmaP{k}}$-complete, if the nesting depth of $\phi$ and
$\psi$ is bounded by a constant $k\geq 0$.

For model checking, things are slightly different. Under a common
bitmap representation, in which $M\models
a$ for any ground atom $a$ is represented by a designated bit, the complexity of
model checking in $\Lcirc$ does not increase, since the (exponential)
size of the explicitly given model $M$ compensates the succinctness of
implicit representation. 

\begin{theorem}
Model checking for first-order $\Lcirc$  under DCA and UNA is \PSPACE-complete.     
\end{theorem}
Notice, however, that the problem is \PSPACE-hard already for
sentences of $\circum$-nesting depth 0, i.e., for ordinary first-order
sentences, since model checking for a given first-order sentence is
\PSPACE-hard.  As easily seen, model checking for first-order NATs is
also \PSPACE-complete, if the arities of abnormality predicates used do not
exceed the arities of the other predicates and the functions by a
constant factor, which is expected to be the case in practice. Similar
as for $\Lcirc$, the problem is \PSPACE-hard already for NATs of
nesting depth 0. However, in the general case, the complexity can be
seen to increase beyond NEXP; we leave a detailed investigation of
this for further work.
%

\section{Comparison to Other Generalizations of Circumscription}
\label{sec:comparison}
\label{sec:discuss}

In this section, we briefly compare nested circumscription to some
other generalizations of circumscription from the literature, namely
prioritized circumscription \cite{lifs-85a,lifs-94} and theory curbing
\cite{eite-etal-93}. Although there are several other generalizations, cf.\
\cite{lifs-94}, the ones considered here are of particular interest
since the former has close semantic relationships to nested
circumscription, while the latter is similar in terms of the complexity.

\subsection{Prioritized Circumscription}

Prioritized circumscription \cite{lifs-85a,lifs-94} generalizes
circumscription $\circum(\phi;P;Z)$ by partitioning the letters $P$
into priority levels $P_1 >$$P_2 >\cdots > P_n$; informally it
prunes all models of $\phi$ which are not minimal on $P_i$, while
$Z\cup P_{i+1}\cup\cdots\cup P_n$ floats and $P_1\cup\cdots\cup
P_{i-1}$ is fixed, for $i=1,\ldots, n$ (cf.\ \cite{lifs-94}).  This
can be readily expressed as the nested circumscription $\psi_n$,
where 
\begin{eqnarray*}
\psi_1 &=&   \circum(\phi;P_1;Z\cup P_2\cup\cdots\cup P_n), \\
\psi_i &=& \circum(\psi_{i-1};P_i;Z\cup P_{i+1}\cup\cdots\cup
P_n),\quad i= 2,\ldots, n.
\end{eqnarray*}
Thus, prioritized circumscription is semantically subsumed by
$\Lcirc$.  Compared to ordinary circumscription, the complexity does
not increase, as inference and model checking remain
$\PiP{2}$-complete and $\CONP$-complete, respectively. 

Intuitively, the reason is that prioritization allows only for a
restricted change of the role of the same letter in iterations (from
floating to minimized and from minimized to fixed), which forbids to
reconsider the value of minimized letters at a later stage of
minimization. This enables a characterization of the models of a
prioritized circumscription as the minimal models of a preference
relation $ \leq_{P_1,\ldots, P_n;Z}$ on the models, where $M
\leq_{P_1,\ldots, P_n;Z} M'$ holds if and only if $M$ and $M'$ coincide on
the fixed letters and either $M$ and $M'$ coincide on all $P_i$, or
$M$ is smaller than $M'$ on the first $P_i$ on which $M$ and $M'$ are
different. This preference relation is polynomial-time computable.  On
the other hand, $\Lcirc$ formulas (and similarly NATs) permit
that minimized letters are reconsidered at a later stage, by making
them floating. This prevents a simple, hierarchical preference
relation as the one for prioritized circumscription.

%


\subsection{Theory Curbing}

Theory curbing is yet another extension of circumscription
\cite{eite-etal-93,eite-gott-00}. Rather than the (hierarchical) use
of circumscription applied to blocks, curbing aims at softening
minimization, and allows for inclusive interpretation of disjunction
where ordinary circumscription returns exclusive
disjunction. Semantically, $\curb(\phi;P;Z)$ for a formula $\phi\in
\Lang$ is the smallest set $\M \subseteq \mod(\phi)$ which contains
all models of $\circum(\phi;P;Z)$ and is closed under minimal
upper bounds in $\mod(\phi)$. A {\em minimal upper bound} (mub) of a set
$\M'$ of models in $\mod(\phi)$ is a model $M \in \mod(\phi)$ such
that (1) $M' \leq_{P;Z} M$, for every $M'\in \M'$, and (2) there
exists no $N\in \mod(\phi)$ satisfying item~1 such that $N<_{P;Z} M'$.

\begin{example}\label{exa:hnp}
{\rm Suppose Alice is in a room with a painting, which she hangs on
the wall $p$ if he has a hammer ($h$) and a nail ($n$). It is
known that Alice has a hammer or a nail or both. This scenario is
represented by the formula $\phi$ in Figure~\ref{fig:hnp}. The models
of $\phi$ are marked with bullets; the desired
models are $\{h\}$, $\{n\}$, and $\{h,n,p\}$, which are
encircled. Circumscribing $\phi$ by minimizing all letters, i.e.,
$\circum(\phi;\{h,n,p\};\emptyset)$ yields the two minimal
models $\{h\}$ and $\{n\}$ (see Figure~\ref{fig:hnp}).
\begin{figure}[bht]
\begin{center}
\bigskip

\setlength{\unitlength}{0.0075in}%
\begin{picture}(320,118)(163,653)
\thinlines
\put(340,690){
\begin{minipage}{0cm}
\mbox{$\phi \;=\;(h \lor n) \land ((h\land n) \limplies p)$}
\end{minipage}}
\put(225,750){\circle{10}}
\put(175,679){\circle*{6}}
\put(175,679){\circle{10}}
\put(275,720){\circle*{6}}
\put(225,690){\circle{10}}
\put(225,690){\circle*{6}}
\put(225,710){\circle*{6}}
\put(225,750){\circle*{6}}
\put(225,690){\circle*{6}}
\put(225,710){\circle*{6}}
\put(225,750){\circle*{6}}
\put(225,750){\line( 0,-1){ 40}}
\put(226,709){\line( 5,-3){ 49.265}}
\put(175,720){\line( 5,-3){ 48.529}}
\put(175,720){\line( 5, 3){ 48.824}}
\put(225,649){\line( 5, 3){ 50}}
\put(275,720){\line( 0,-1){ 41}}
\put(225,690){\line( 0,-1){ 41}}
\put(225,690){\line( 5, 3){ 50}}
\put(175,720){\line( 0,-1){ 40}}
\put(225,751){\line( 5,-3){ 50}}
\put(275,720){\line( 0,-1){ 40}}
\put(225,690){\line( 5, 3){ 50}}
\put(225,750){\line( 0,-1){ 40}}
\put(174,679){\line( 5, 3){ 50}}
\put(175,720){\line( 0,-1){ 40}}
\put(225,690){\line( 0,-1){ 40}}
\put(174,679){\line( 5,-3){ 50}}
\put(225,751){\line( 5,-3){ 50}}
\put(285,675){\makebox(0,0)[lb]{\raisebox{0pt}[0pt][0pt]{$\{p\}$}}}
\put(225,758){\makebox(0,0)[b]{\raisebox{0pt}[0pt][0pt]{$\{h,n,p\}$}}}
\put(225,670){\makebox(0,0)[lb]{\raisebox{0pt}[0pt][0pt]{$\{n\}$}}}
\put(230,720){\makebox(0,0)[b]{\raisebox{0pt}[0pt][0pt]{$\{h,p\}$}}}
\put(285,720){\makebox(0,0)[lb]{\raisebox{0pt}[0pt][0pt]{$\{n,p\}$}}}
\put(165,720){\makebox(0,0)[rb]{\raisebox{0pt}[0pt][0pt]{$\{h,n\}$}}}
\put(225,633){\makebox(0,0)[b]{\raisebox{0pt}[0pt][0pt]{$\emptyset$}}}
\put(165,675){\makebox(0,0)[rb]{\raisebox{0pt}[0pt][0pt]{ $\{h\}$}}}
\end{picture}
\end{center}
\caption{
\label{fig:hnp}
\protect{The hammer-nail-painting example}
}
\end{figure}
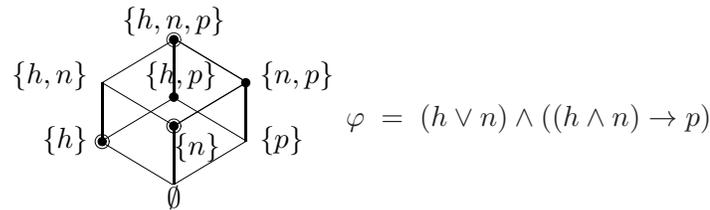
Since $p$ is false in the minimal models, circumscription tells us
that Alice does not hang the painting up. One might argue that $p$
should not be minimized but fixed under circumscription.  However,
starting with the model of $\phi$ where $h,n$ and $p$ are all
true and then circumscribing with respect to $h$ and $n$ while
keeping $p$ true, we obtain the smaller models $\{h,p\}$ and
$\{n,p\}$, which are not very intuitive. The remaining possibility
is to let $p$ float. However, this does not work either, since the
circumscription $\circum(\phi;h,n;p) \equiv ((h \leftrightarrow \neg n) \land \neg p)$ is equivalent to
$\circum(\phi;h,n,p;\emptyset)$
On the other hand, the model
$\{h,n,p\}$, which corresponds to the inclusive interpretation
of the disjunction $h\lor n$, seems plausible. Under curbing, we
obtain the desired models from $\curb(\phi;h,n;p)$.
\hfill$\Box$}
\end{example}

Like for $\Lcirc$ and NATs, inference and model checking for
$\curb(\phi;P;Z)$ are \PSPACE-complete \cite{eite-gott-00} in the
propositional context, and can be shown to have likewise exponentially
higher complexity in the datalog setting (i.e., in a function-free
language under DCA and UNA, cf.\ Section~\ref{sec:fo}).

However, while the complexity is the same, curbing and NATs have
different expressiveness, if we consider these formalisms as
query languages for uniformly expressing properties over collections
of ground facts, such as 3-colorability of graphs which are described
by their edge relations. It turns out that curbing can express some
properties which $\Lcirc$ and NATs (most likely) can not express. For
example, we can write a (fixed) interpreter $T_I$ in this language for
curbing {\em varying} propositional 3CNF formulas $\phi$, input as
ground facts $F(\phi)$, such that the curb models of $T_I \cup
F(\phi)$ and of $\phi$ are in 1-1 correspondence.  Notice that curbing
such 3CNFs $\phi$ is \PSPACE-complete,
and thus, by well-known results in complexity, this is not expressible
by any fixed $\Lcirc$ formula or NAT (unless PH=\PSPACE).

We elaborate on this interpreter for propositional curbing in more
detail. The constants represent the propositional atoms, and the
clauses of $\phi$ are stored using 3-ary predicates $R_0,R_1,R_2$, and
$R_3$, where $R_i(x_1,x_2,x_3)$ intuitively represents the clause
$\bigvee_{j=1}^ix_j\lor \bigvee_{j=i+1}^3 \neg x_j$. E.g.,
$R_2(a,c,b)$ represents the clause $a\lor c \lor \neg b$. Unary
predicates {\it pvar} and {\it zvar} are used for designating the
atoms in $P$ and $Z$, respectively.

The theory $T_I$ is as follows:
$$
\renewcommand{\arraystretch}{1.25}
\begin{array}{r@{\,}l@{\,}r@{\,\lor\,}c@{\,\lor\!\!\!\!\!\!}c}
\forall x,y,z.& R_0(x,y,z) \limplies & \neg t(x) & \neg t(y) &\neg t(z),\\ 
\forall x,y,z.& R_1(x,y,z) \limplies &      t(x) & \neg t(y) &\neg t(z),\\ 
\forall x,y,z.& R_2(x,y,z) \limplies &      t(x) &      t(y) &\neg t(z),\\ 
\forall x,y,z.& R_3(x,y,z) \limplies &      t(x) &      t(y) & t(z),\\ 
\forall x.& \multicolumn{4}{@{\,}l}{p(x) \lequiv (\mathit{pvar}(x) \land t(x)),} \\
\forall x.& \multicolumn{4}{@{\,}l}{q(x) \lequiv ( \neg \mathit{pvar}(x)\land \neg \mathit{zvar}(x) \land t(x)).}
\end{array}
$$
Intuitively, $t(x)$ means that $x$ has value true. Here, the predicate
$p$ is minimized, while $q$ is fixed and $t$ is floating.

The set of facts $F(\phi)$ contains 
\begin{enumerate}
\item for each clause $(\neg)a
\lor (\neg)b \lor (\neg)c$ from $\phi$ the respective atom
$R_i(a,b,c)$; 
\item for each $p\in P$ (resp., $z\in Z$) the atom $\mathit{pvar}(p)$
(resp., $\mathit{zvar}(z)$); 
\item the negations of all other ground
atoms (i.e., $F(\phi)$ is the CWA given the atoms in 1 and 2).
\end{enumerate}

\begin{example}\label{exa:hnp-contd}
{\rm Reconsider $\curb(\phi;h,n;p)$ for the formula $\phi=(h
\lor n)\land (\neg h\lor \neg n \lor p)$ (rewritten as a CNF)
from Example~\ref{exa:hnp}. Then, the constants are $h,n,p$. The
positive facts in $F(\phi)$ are $R_3(h,n,n)$ and $R_1(p,h,n)$
encoding the first and the second clause of $\phi$, respectively
(where we add a redundant disjunct $n$ in the first clause), and 
$\mathit{pvar}(h)$, $\mathit{pvar}(n)$, and $\mathit{zvar}(p)$.

Note that $T_I \cup F(\phi)$ logically implies $t(h) \lor t(n)$,  
$t(p) \lor \neg t(h) \lor \neg t(n)$, $\neg p(p)$, $p(h) \lequiv
t(h)$, $p(n) \lequiv t(n)$, $\neg q(h)$, $\neg q(n)$, and $\neg q(p)$.
Thus, Herbrand models of $T_I\cup F(\phi)$ may differ only on the
atoms $t(h)$, $t(n)$, $t(p)$, $p(h)$, and $p(n)$. The feasible assignments
of these atoms correspond to the models of $\phi$. If $M$ is a
model of $\phi$, then by assigning true to the atoms $t(a)$ where $a \in M$
and $p(a)$ where $a \in M \cap \{h,n\}$, we obtain a feasible such
truth assignment. On the other hand, if $M$ is a Herbrand model of $\curb T_I \cup
F(\phi);p;t)$, then $\{ a \mid t(a)\in M\}$ is a model of
$\phi$. Overall, the Herbrand models of $T_I \cup F(\phi)$
correspond 1-1 to the models of $\phi$.}
\hfill$\Box$
\end{example}

The following proposition, whose proof is omitted, states that the
interpreter works similarly in the general case. 

\begin{proposition}
Under DCA and UNA, the models of $\curb(T_I \cup F(\phi);p;t)$ and 
$\curb(\phi;P;Z)$ are in 1-1 correspondence. 
\end{proposition}

 From results in \cite{eite-gott-00}, we easily obtain that evaluating
any given QBF $\Phi$ (which is \PSPACE-complete) is polynomially
reducible to deciding $\curb(\phi,P;Z)\models \neg a$, where $\phi \in
\Lang$ is in 3CNF and $a$ is an atom. Thus, $\curb(T_I\cup
F(\phi));p;t)\models \neg t(a)$ expresses evaluating the QBF $\Phi$
given by $F(\phi)$.

On the other hand, unless PH = \PSPACE, a ``datalog''
$\Lcirc$ formula resp.\ NAT similar to $T_I$ does not exist: due to fixed
nesting depth, it can only express a problem in PH. 

Further relationships between $\Lcirc$ resp.\ NATs and curbing, as
well as other expressive knowledge representation formalisms 
(e.g., \cite{bara-etal-00,eite-etal-97g,poll-remm-97}), remain to be
explored.

\section{Conclusion}
\label{sec:conclusion}

In this paper, we have studied the computational complexity of the
logical language $\Lcirc$, which is a propositional language that
allows the nested use of circumscription, and of the propositional
fragment of nested abnormality theories (NATs) that were proposed by
Lifschitz \cite{lifs-95} as an elegant circumscriptive framework for
modularized knowledge representation.  As we have shown, NATs can be
regarded as a semantic fragment of $\Lcirc$. As it turned out, NATs
and thus $\Lcirc$ are capable of expressing more difficult (in terms
of complexity) problems than ordinary unnested circumscription, and
can represent \PSPACE-complete problems. Furthermore, we have
identified fragments of NATs which have lower complexity, where we
focused on generalizations of Horn CNFs, such as Horn logic programs
and the DATALOG$^{Circ}$ query language \cite{cado-palo-98}. In
particular, we have provided an efficiently computable normal form for
nested logic programs.  Finally, we have compared nested
circumscription to other generalizations of circumscription.

Our results give a clear picture of the complexity situation, and
reveal nesting and the use of local variables in NATs as sources of
complexity. This gives useful insight into the complexity of $\Lcirc$
formulas and NATs, which is useful for understanding their
computational nature and requirements. For example, it can be
fruitfully exploited in considerations on eliminating nestings, or on
changes to the set of defined letters in a NAT. To give a concrete
example, suppose we have an extended Horn NAT $\T$ which has nesting
depth one. Then, by Theorem~\ref{theo:xhorn-mc+inf}, inference of a
formula $\phi$ from $\T$ is $\PiP{2}$-complete in general, and thus
can be polynomially transformed to a standard circumscriptive theorem
prover. If, moreover, the blocks inside $\T$ have no fixed letters and
do not use $\max$, then by Theorem~\ref{theo:horn-nat-no-fixed} we can
efficiently eliminate nesting from $\T$, and transform inference
$\T\models \phi$ via a standard Horn circumscription to a
SAT solver in polynomial time. 

While we have addressed and resolved the main issues concerning the
complexity of nested circumscription in a propositional setting in
this paper, several issues remain for future work:

\begin{myitemize}

\item On the complexity side, our study may be extended to cover
further fragments of NATs and $\Lcirc$ besides the ones considered in
this paper. Besides Horn theories, other syntactic fragments were
e.g.\ considered in \cite{cado-lenz-94}, which provides a good starting point
for such a programme. Furthermore, a detailed study of the complexity
of nested circumscription in the first-order case and restricted
fragments (monadic theories, etc) would be interesting.

\item Complementing the results on reasoning complexity, Cadoli {\em
et al.} \cite{cado-etal-96,cado-etal-00}, Gogic {\em et al.}
\cite{gogi-etal-95}, Selman and Kautz \cite{selm-kaut-96}, Darwiche
and Marquis \cite{darw-marq-01,darw-01} and others have studied
representability issues among KR formalisms,
considering problems like representing theories in one KR formalism
with polynomial resources in another target formalism, such that
the set of models or certain inference relations are preserved. In
particular, ``knowledge compilation,'' whose idea is that off-line
preprocessing with high computational resources might help to speed up
on-line reasoning, and make sometimes intractable problems tractable,
has been attracting attention during the last years (see
\cite{cado-doni-97} for an initial survey). A study of representation
and compilability aspects of $\Lcirc$ and NATs, and a comparison to
other KR formalisms remains as an interesting issue.
In particular, it would be interesting to determine under which circumstances
NATs can be compiled in other NATs with lower nesting.

\item An important instance of the issue in the previous paragraph is
when a NAT can be efficiently replaced by an equivalent standard
or prioritized circumscription, or even by an ordinary
propositional formula. Notice that this issue is highly significant
for algorithms that implement NATs on top of circumscriptive theorem
provers or classical SAT solvers. Our results give a very preliminary
answer to this question, by showing that this is, e.g., possible for
Horn NATs without fixed letters. However, other and more expressive
fragments might be identified which have this property.

\item Finally, it remains to develop efficient algorithms and methods
for computing NATs, either by reduction to an engine for some related
KR formalism or logic, or by designing genuine algorithms. Su's CS
program \cite{su-95} and Doherty {\em et al.}'s DLS algorithm
\cite{dohe-etal-97,gust-96}, which handle the case of predicate logic,
are incomplete in general and presumably not highly efficient in the
propositional context. The use of QBF solvers (e.g.,
\cite{cado-etal-02,rint-99,feld-etal-00}) is here a suggestive
starting point for obtaining more suitable systems.
\end{myitemize}

As we believe, addressing these issues is worthwhile since nesting
circumscriptions is a natural generalization of circumscription, and
yields, as shown by our results, a simple yet expressive knowledge
representation formalism for encoding reasoning tasks with complexity
in PSPACE.

\subsubsection*{Acknowledgments}

This work was supported by the Austrian Science Fund (FWF) Project
Z29-INF. We are grateful to the referees of the preliminary conference
version of this paper, which had a number of useful suggestions for
improvements.

\newcommand{\bibpath}{/home/staff/eiter/tex/bibs}

{\small 

\bibliographystyle{abbrv}

\ifmakebbl

\bibliography{\bibpath/eiterlib+,\bibpath/survey,\bibpath/library,/home/staff/eiter/papers/bibtex/bibtex,nat}

\else

\newcommand{\SortNoOp}[1]{}

\fi 

}

\end{document}